\def\year{2025}\relax
\documentclass[letterpaper]{article} 
\usepackage{aaai22}  
\usepackage{times}  
\usepackage{helvet}  
\usepackage{courier}  
\usepackage[hyphens]{url}  
\usepackage{graphicx} 
\usepackage{natbib}  
\usepackage{caption} 
\DeclareCaptionStyle{ruled}{labelfont=normalfont,labelsep=colon,strut=off} 
\usepackage{algorithm}
\usepackage[noend]{algpseudocode}
\usepackage[table]{xcolor}
\definecolor{orange}{rgb}{1,0.71,0.439}
\usepackage{enumitem} 

\usepackage{xcolor}

\usepackage{newfloat}
\usepackage{listings}
\floatstyle{ruled}
\newfloat{listing}{tb}{lst}{}
\floatname{listing}{Listing}

\usepackage{amsmath,amsfonts}
\usepackage{multirow}
\usepackage{array}

\title{Masculine Defaults via Gendered Discourse in Podcasts and\\Large Language Models}
\author {
    Maria Teleki,
    Xiangjue Dong,
    Haoran Liu,
    James Caverlee
}
\affiliations {
    Texas A\&M University\\
    \{mariateleki, xj.dong, liuhr99, caverlee\}@tamu.edu
}

\begin{document}

\maketitle

\begin{abstract}
Masculine defaults are widely recognized as a significant type of gender bias, but they are often unseen as they are under-researched. Masculine defaults involve three key parts: (i) the cultural context, (ii) the masculine characteristics or behaviors, and (iii) the reward for, or simply acceptance of, those masculine characteristics or behaviors. In this work, we study \textit{discourse-based masculine defaults}, and propose a twofold framework for (i) the large-scale discovery and analysis of gendered discourse words in spoken content via our \textit{Gendered Discourse Correlation Framework (GDCF)}; and (ii) the measurement of the gender bias associated with these gendered discourse words in LLMs via our \textit{Discourse Word-Embedding Association Test (D-WEAT)}. We focus our study on podcasts, a popular and growing form of social media, analyzing 15,117 podcast episodes. We analyze correlations between gender and discourse words -- discovered via LDA and BERTopic -- to automatically form gendered discourse word lists. We then study the prevalence of these gendered discourse words in domain-specific contexts, and find that gendered discourse-based masculine defaults exist in the domains of business, technology/politics, and video games. Next, we study the representation of these gendered discourse words from a state-of-the-art LLM embedding model from OpenAI, and find that the masculine discourse words have a more stable and robust representation than the feminine discourse words, which may result in better system performance on downstream tasks for men. Hence, men are rewarded for their discourse patterns with better system performance by one of the state-of-the-art language models -- and this embedding disparity is a representational harm and a masculine default.

\end{abstract}

\raisebox{-0.38\height}{\includegraphics[height=0.75cm]{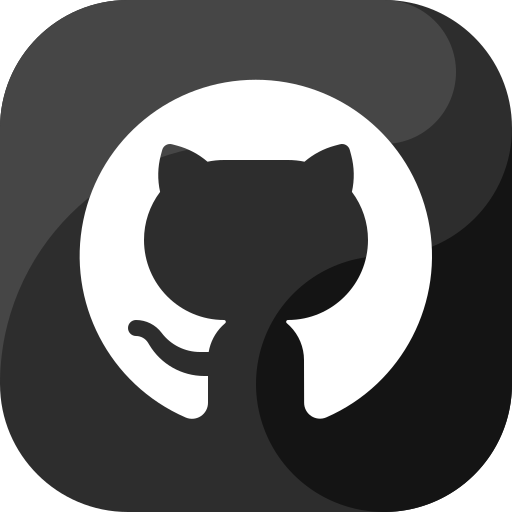}}
{\fontsize{7.5pt}{6pt} \space \texttt{github.com/mariateleki/masculine-defaults}}

\section{Introduction}

\begin{figure}[t]
\centering \includegraphics[scale=0.09]{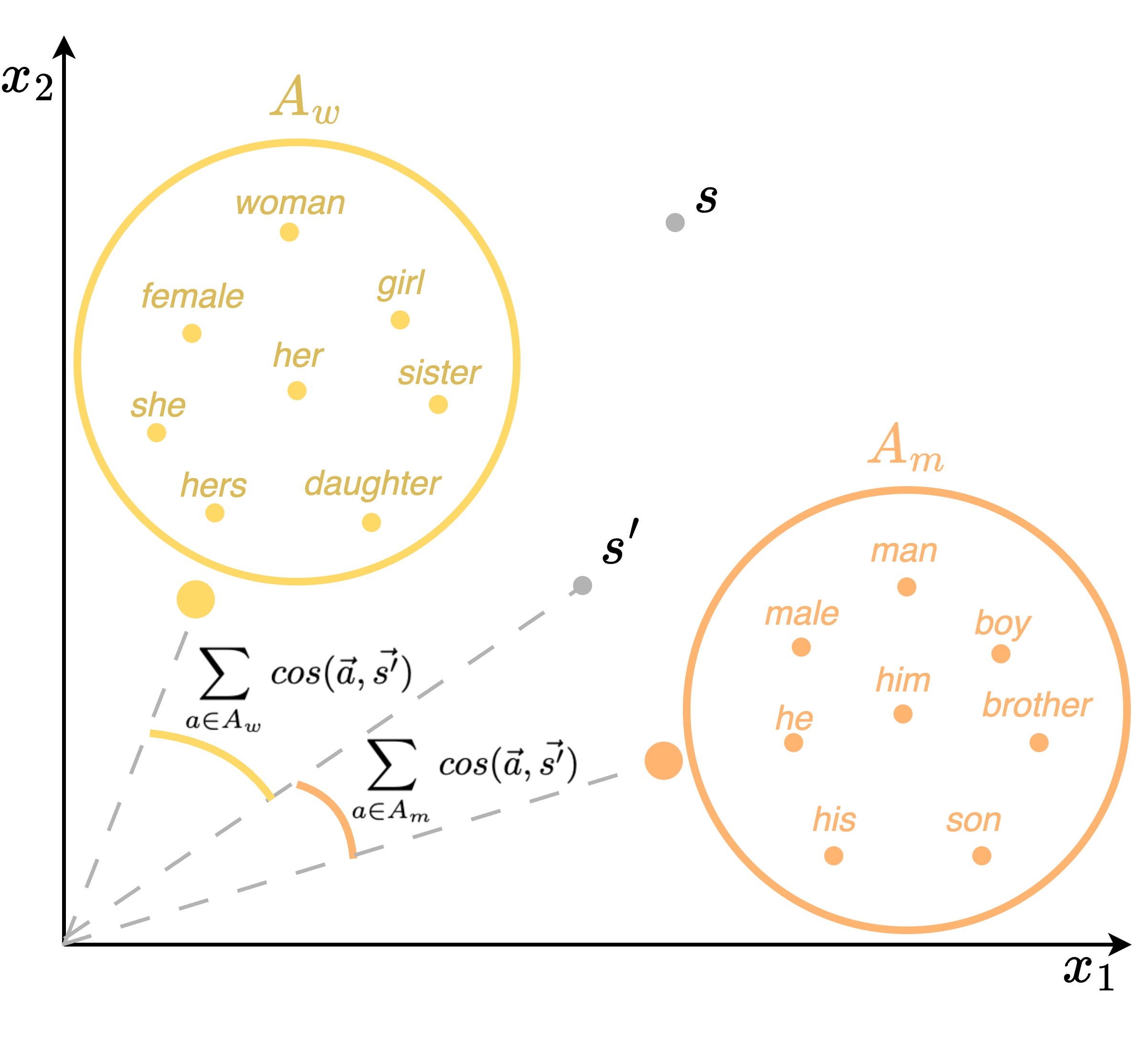} \caption{An overview of our two-part framework: (i) Using our Gendered Discourse Correlation Framework (GDCF, as shown in Figure \ref{fig:BigDiagram}), we obtain \textit{gendered discourse word lists}. (ii) We then perform our Discourse Word-Embedding Association Test (D-WEAT, as shown here in Figure \ref{fig:embds}). We form parallel sentences, $s$ and $s'$, by swapping masculine discourse words (e.g. ``going'') for feminine discourse words (e.g. ``like''): $s\!=\,$\textit{And I was \textbf{going}, hey, it's cold outside...}, and $s'\!=\,$\textit{And I was \textbf{like}, hey, it's cold outside...} We find that the masculine discourse words have a more stable embedding representation -- this is a representational harm and a masculine default.}
\label{fig:embds}
\end{figure}

\textit{Masculine defaults} are a type of gender bias ``in which characteristics and behaviors associated with the male gender role are valued, rewarded, or regarded as standard, normal, neutral, or necessary aspects of a given cultural context'' \cite{Cheryan.2020}. Hence, there are three parts to a masculine default: (i) the cultural context, (ii) the male characteristics or behaviors, and (iii) the reward for, or simply acceptance (neutral) of the male characteristics or behaviors. 

\textbf{Hence, to determine whether or not a behavior constitutes a masculine default \cite{Cheryan.2020}, we consider: What is the reward or standard associated with a given masculine behavior or characteristic?} For example, in the cultural context of the United States, the prevalence of men in computer science is a masculine default, as men are economically rewarded for being computer scientists via statistically higher salaries \cite{wagestats}, and are largely socially accepted in this role \cite{Cheryan.2020}. This masculine default, then, propagates social injustice, as ``women feel a lower sense of belonging and anticipate less success'' in computer science, and do not enter the field at a comparable rate to men and reap the economic rewards \cite{Cheryan.2020, degreestats}. These masculine defaults result in the \textit{other-ing} of women \cite{Beauvoir.1949}.

Considerable prior research has examined gender differences in social media (e.g., \citet{wang2021representation, kalhor2023gender, johnson2021global, wang2019gender}) and in LLMs (e.g., \citet{dong-etal-2023-co2pt, Caliskan.2017, May.2019, Bolukbasi.2016}). But how do \textit{masculine defaults} manifest on social media? And how do they impact emerging systems like large language models (LLMs) that are trained in part over social media?  While masculine defaults are highly connected to gender differences,\footnote{We consider the binary definitions of sex (female/male) and gender (women/men, feminine/masculine) in our work due to (i) continuity with previous work in the gender debiasing task in the NLP community \cite{Caliskan.2017, Bolukbasi.2016}, and (ii) modeling constraints -- i.e., we use \textit{inaSpeechSegmenter} \cite{ddoukhanicassp2018} for gender approximation via the podcast audio signal. The binary gender definition, however, is not representative of the sex and gender spectrums, and transgender, intersex, intersectional identities, and other identities are also not represented \cite{ghai2021wordbias, Ovalle_Goyal_Dhamala_Jaggers_Chang_Galstyan_Zemel_Gupta_2023, Seaborn_Chandra_Fabre_2023}. This is an important direction for future work.} there is a research gap in identifying and analyzing masculine defaults that arise through \textit{gender differences in discourse}. 

Specifically, we focus on patterns of discourse in spoken communication, including fillers (e.g., \textit{uh}, \textit{um}), discourse markers (e.g., \textit{well}, \textit{you know}, \textit{I mean}), false starts (e.g., \textit{It was, anyways, I went to Target yesterday}) and more \cite{merriam_2024, shriberg1994preliminaries}. Such discourse words are non-content related words that serve important social purposes with respect to gender, such as to \textit{``hold the floor''} in conversation \cite{shriberg1994preliminaries, shriberg1996disfluencies}. Previous work notes gender differences in how men and women use specific types of \textit{discourse words} -- for example, men use more filled pauses and repeats \cite{shriberg1996disfluencies, bortfeld2001disfluency} than women. However, these studies lack an automated method for large-scale discourse word discovery and gender analysis, primarily relying on the Switchboard corpus \cite{mitchell1999treebank} -- an older, human-annotated corpus which is not representative of the range of natural speech patterns, as the phone calls were recorded in the manufactured, awkward situation of randomly-pairing two callers and assigning them a topic to discuss. 

Hence, we propose in this paper a twofold framework for (i) the large-scale discovery and analysis of gendered discourse words in spoken content via our \textit{Gendered Discourse Correlation Framework (GDCF)}; and (ii) the measurement of the gender bias associated with these gendered discourse words in LLMs via our \textit{Discourse Word-Embedding Association Test (D-WEAT)}. Concretely, we focus our study on podcasts, a popular and growing form of social media~\cite{clifton2020}. According to Pew Research, ``42\% of Americans ages 12 and older have listened to a podcast in the past month'' as of 2023 compared to 12\% in 2013~\cite{audio_podcasting_fact_sheet}. We analyze the \textit{rewards} associated with \textit{gendered discourse words} in 15,117 podcast episodes from the Spotify Podcast Dataset \cite{clifton2020} -- i.e., discourse words with significant positive correlations with either men or women --  to determine whether or not masculine defaults are present. Our study is organized around the following research questions:

\begin{itemize}
\item \textit{RQ0: How are women and men's discourse different?}
\item \textit{RQ1: Are discourse-based masculine defaults present in domain-specific contexts?}
\item \textit{RQ2: Are discourse-based masculine defaults present in LLM embeddings?} 
\end{itemize}

We first (RQ0) introduce our \textit{Gendered Discourse Correlation Framework (GDCF)}, a framework for discovering gendered discourse words, with features which are centered around spoken content -- specifically, an audio-based \textsc{Gender Segmenter} \cite{ddoukhanicassp2018}, a \textsc{Topic Modeler} via LDA \cite{blei2003latent} and BERTopic \cite{grootendorst2022bertopic}, and a specialized \textsc{Conversational Parser} \cite{jamshid-lou-johnson-2020-improving}. We analyze correlations between \textit{gender} and \textit{discourse words} to automatically form gendered discourse word lists. 
Additionally, GDCF is a flexible framework which can be extended to other forms of audio speech data -- such as short videos that are prevalent on TikTok, Instagram, and YouTube, long videos on YouTube, streamers on Twitch, and more.

We then study (RQ1) the prevalence of these gendered discourse words in domain-specific contexts. We find that masculine discourse words are positively correlated with the business domain. Because participation in the business domain grants economic rewards, there are indeed discourse-based masculine defaults present in the business domain. We additionally show that this is the case for the domains of technology/politics and video games, and provide more, related results in the Appendix.

Next, we study (RQ2) the representation of these gendered discourse words in a state-of-the-art LLM embeddings model from OpenAI, {\fontfamily{qcr}\selectfont text-embedding-3-large}. We find that the masculine discourse words have a more stable and robust representation than the feminine discourse words, resulting in better system performance on downstream tasks for men. Hence, men are rewarded for their discourse patterns with better system performance by one of the state-of-the-art language models -- and therefore this difference in the embedding representations for women and men constitutes a \textit{representational harm} \cite{Blodgett.2020} and a masculine default. 

\textbf{Resources.} We release our code at \url{https://github.com/mariateleki/masculine-defaults} and the extended results at \url{https://www.gendered-discourse.net}.

\section{Related Work}

\subsubsection{Sex, Gender, and Language.}
We focus in our work on \textit{gender} rather than \textit{sex}:$^1$ sex~(female/male) is established based on biology; whereas, gender~(women/men, feminine/masculine) ``is the activity of managing situated conduct in light of normative conceptions of attitudes and activities appropriate for one's sex category'' \cite{West.1987,Unger.1979,Muehlenhard.2011}. Gender is something that people ``do,'' and ``gender [can be understood] as a routine, methodological, and recurring accomplishment'' \cite{West.1987}. In Butler's theory of \textit{gender performativity}, ``[g]ender is instituted through the stylization of the body and, hence, must be understood as the mundane way in which bodily gestures, movements, and enactments of various kinds constitute the illusion of an abiding gendered self'' \cite{butler1988performative} -- an \textit{enactment}, then, includes \textit{language}: the way women and men speak. Butler argues that conforming to this \textit{gender schema}  -- wherein certain ``attitudes,'' ``activities,'' ``attributes,'' and ``behaviors'', including language, are assigned to either women or men \cite{Bem.1984, West.1987} -- is necessary for women to ``ask for recognition in the law or in political life'' \cite{butler2009performativity}. In this way, masculinities and femininities relate to social and political power.

\textit{Hegemonic masculinity} refers to a performative, ```currently accepted' strategy'' for maintaining the patriarchal imbalance of social and political power via cultural dominance \cite{connell1995masculinities, connell1987gender}. Maintaining power necessitates different strategies over time, thus, hegemonic masculinity is highly contextual. One recent type of hegemonic masculinity is \textit{technomasculinity} -- the form of masculinity associated with high-tech professions, such as engineering and science \cite{Cooper_2000, lockhart2015nerd, Bulut+2020, goree2023really}. Hence, as gender and gender roles are highly contextual, we limit our definition of gender temporally, to recently, and geographically, to the United States.

Hegemonic masculinity, then, is closely related to \textit{masculine defaults}, which are a form of \textit{other-ing} -- conciously and/or subconciously -- that occurs as the result of the masculine social and political hierarchy. ``Masculine defaults include ideas, values, policies, practices, interaction styles, norms, artifacts, and beliefs that often do not appear to discriminate by gender but result in disadvantaging more women than men'' \cite{Cheryan.2020}. Masculine defaults relate to other-ing in that ``alterity is the fundamental category of human thought'' and ``He is the Subject; he is the Absolute. She is the Other'' \cite{Beauvoir.1949}. In other words, he is the \textit{default}, and she is the other. An example of a masculine default in language is the use of \textit{masculine generics}, as ``[a]n almost universal and fundamental asymmetry lies in the use of masculine generics. In English, for example, generic he can be used when gender is irrelevant (e.g., the user... he)'' rather than `she' \cite{Sczesny.2016}. For a masculine behavior to be considered a \textit{masculine default}, there must be a \textit{reward} or a \textit{standard} associated with the use of the masculine behavior \cite{Cheryan.2020}.

\subsubsection{Podcast Language Analysis.} 

Podcasts have come under increased research scrutiny in the past few years. For example,  \citet{yang2019more} analyzed non-textual characteristics of podcasts (like energy or seriousness) through audio spectrogram representation learning methods. \citet{clifton2020} conducted an analysis of the Spotify dataset podcasts, where they also found that discourse topics exist, and they found a higher frequency of first-person pronouns and amplifiers as compared to the Brown corpus. \citet{Valero2022} studied topic modeling on podcasts for information retrieval with the Spotify dataset. \citet{martikainen2022exploring} have examined how stylistic features relate to genres on a small scale using PCA and k-means clustering: they analyzed a subset of 14 episodes then a subset of 911 episodes. They also used \textit{inaSpeechSegmenter}~\cite{ddoukhanicassp2018} to obtain gender correlations. \citet{rezapour2020spotify} looked at using the iTunes topics and named entities to generate extractive summaries.

Closest to this work, \citet{reddy-etal-2021-modeling} analyzed the relationships between linguistic features and engagement (measured via podcast popularity) over the Spotify dataset. In contrast, our work focuses on measuring feature correlations related to \textit{gender} and \textit{discourse}. Hence, we introduce two new modules, the \textsc{Gender Segmenter} module and the \textsc{Conversational Parser} module, to assist us in our aim of focusing on \textit{gender} and \textit{discourse}, rather than popularity.

\subsubsection{Large Language Models (LLM) and Discourse Words.}

\begin{figure*}[t] 
\centering \includegraphics[scale=0.26]{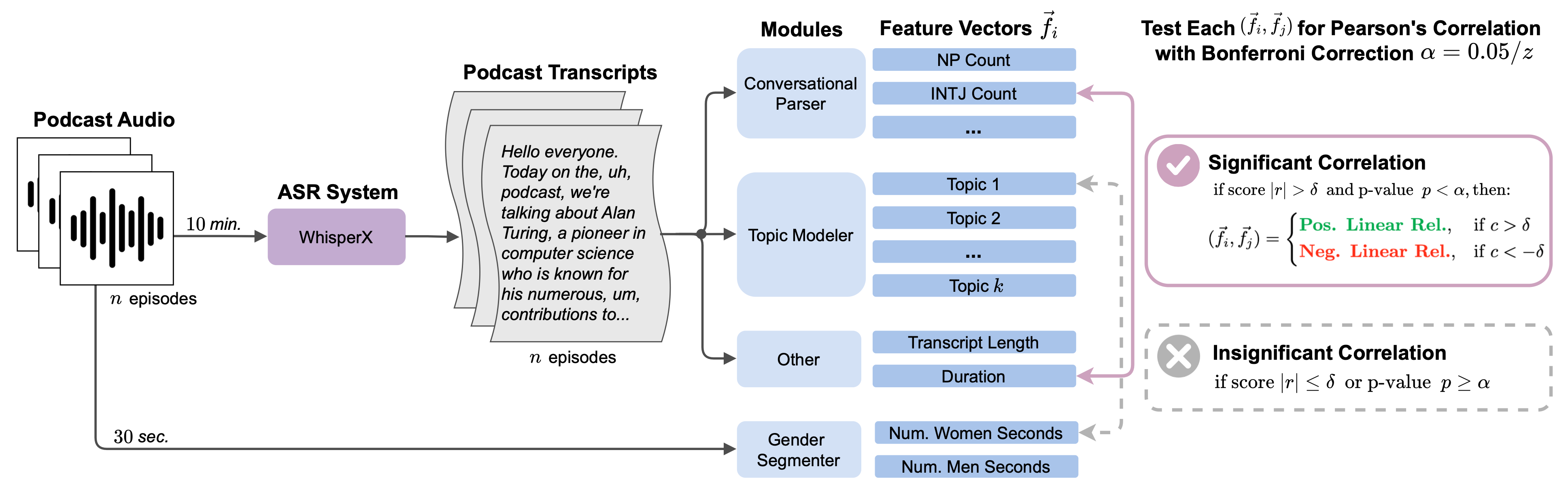} 
\caption{GDCF (Gendered Discourse Correlation
Framework) Diagram: Testing for correlations with an example of a significant correlation and an insignificant correlation -- all $(\Vec{f_i},\Vec{f_j})$ pairs are labeled \textit{significant} or \textit{insignificant}. $\lvert \Vec{f_i} \rvert = 15,117$ podcast episodes. $z=\binom{124}{2}=7,626$ correlation tests for the 124 total feature vectors.}
\label{fig:BigDiagram} 
\end{figure*}

Large language models are trained on gender imbalanced patterns of discourse usage -- be it on podcasts, YouTube videos, and/or other social media formats. It is well-recognized that LLMs can inherit and propagate gender stereotypes \cite{Bolukbasi.2016}.
Thus, with respect to gender, important discourse words should be represented equally in the embedding space, just as stereotype words are via gender debiasing methods \cite{Caliskan.2017, May.2019}. Current gender debiasing methods in natural language processing (NLP) typically ignore this prevalent discourse signal, instead focusing on stereotypes, such as occupational stereotypes like \textit{doctor/nurse} \cite{Bolukbasi.2016}, and other stereotype categories like \textit{science/arts} and \textit{career/family} in the Word-Embedding Association Test (WEAT) \cite{Caliskan.2017} and the WEAT extension, the Sentence Encoder Association Test (SEAT) \cite{May.2019}. Both WEAT and SEAT tests are based on the Implicit Association Test (IAT) from the field of psychology \cite{Greenwald.1998}. 

While not all discourse words are gender-stereotyped, some are. Consider some of the words from our study: \textit{going}, \textit{know}, and \textit{things} are not stereotyped. However, consider the word \textit{like}, which women in popular media tend to use often  -- such as in the iconic line from the 2001 hit movie, \textit{Legally Blonde}: \textit{``What, like, it's hard?''} Hence, we differ from WEAT in that we extend WEAT beyond stereotyping, to include discourse words which are correlated with men or women, and hence carry implicit \cite{Seaborn_Chandra_Fabre_2023, Caliskan.2017, Greenwald.1998, Greenwald.1995} gender information and reinforce masculine defaults. 

\section{GDCF: Gendered Discourse Correlation Framework (RQ0, RQ1)}
In this section, we introduce a framework for discovering gendered discourse words as shown in Figure \ref{fig:BigDiagram}, with features centered around spoken content. We then (RQ0) analyze correlations between between \textit{gender} and \textit{discourse words} to automatically form gendered discourse word lists. Next, (RQ1) we analyze correlations between these gendered discourse words and domains represented by our framework features, and find that, indeed, certain domains (technology/politics, business, and video games) do have discourse-based masculine defaults. 

\subsection{Spotify Podcast Dataset}

The Spotify Podcast Dataset is ``[t]he largest corpus of transcribed speech data, from a new and understudied domain'' \cite{clifton2020}, which consists of 105,360 podcast episodes in total. The podcasts are randomly sampled from January 1, 2019 to March 1, 2020, and they are ``all Spotify owned-and-operated,'' and contain personally identifiable information and offensive content~\cite{clifton2020}. Each podcast \textit{show} has many \textit{episodes} -- analogous to TV shows, where each show has potentially multiple episodes. \citet{clifton2020} sampled a large collection of podcasts and then filtered out (i) non-English podcasts based on the metadata tags and \textit{langid.py}, a pre-trained multinomial naive Bayes learner~\cite{lui-baldwin-2011-cross,lui-baldwin-2012-langid}, (ii) non-professional episodes longer than 90 minutes (hence, this dataset may not well represent beginner/non-professional podcast creators); and (iii) episodes comprised of less than 50\% speech.

In our work, we transcribe the podcasts using WhisperX~\cite{bain2022whisperx}, a state-of-the-art method based on OpenAI's transformer-based Whisper ASR model, which was trained on 680,000 hours of labeled audio data~\cite{radford2023robust} and performs well on discourse-style audio data, as shown in the Appendix in Table \ref{tab:whisperx_googleasr}. (See implementation details in the Appendix). Following~\citet{reddy-etal-2021-modeling}, we apply the following filters to the dataset: (i) truncate episodes to 10 minutes to control for duration; (ii) filter out episodes $<$10 minutes in duration; (iii) remove the over 3,500 non-English podcasts identified by WhisperX (see Figure~\ref{fig:nonenglish} in the Appendix) that had bypassed the original Spotify language filter; and (iv) filter out podcasts with less than 10 words.\footnote{25 podcasts have no transcribed words at all, despite using a speech filter~\citet{clifton2020}. Upon inspection, almost all of these podcasts are ASMR podcast episodes. Similarly, many episodes which have less than 10 words are also ASMR episodes.} Finally, (v) to control for the impact of a single podcast show having potentially many episodes, we follow~\citet{reddy-etal-2021-modeling} and only keep one episode per podcast show in our dataset for normalization of the impact of different podcast shows having different numbers of episodes, resulting in 15,117 episodes. We release the correlation results both with and without filter v at: \url{https://www.gendered-discourse.net/extended-results}. 

\subsection{Features Definition and Extraction}

We define and extract features from the Spotify Podcasts using modules -- the \textsc{Conversational Parser}, the \textsc{Topic Modeler}, the \textsc{Gender Segmenter}, and an \textsc{Other} module (for the transcript length and duration features). Our features are selected to help us in our aim of characterizing the differences between discourse patterns and topics for men and women. We focus in RQ0 and RQ1 on our most novel result, which is the discovery and significance of \textit{discourse-related masculine defaults}. Results related to the \textsc{Conversational Parser} and \textsc{Other} modules result in the large-scale confirmation of small-scale  previous studies, and we provide these results in the Appendix.

\subsubsection{\textsc{Topic Modeler Module.}}

We perform topic modeling on the podcast transcripts to obtain high-quality \textit{genre} features, as the provided features in the dataset are noisy. For example, the creator-provided descriptions and iTunes categories may be search engine optimized (SEO) to gain more podcast streams -- i.e., they may include keywords or other content in an effort to get the podcast ranked higher when users search, rather than to most accurately reflect the content of the podcast. We study two topic models -- LDA with non-contextual embeddings, and BERTopic with contextual embeddings: 
\begin{itemize}
    \item \textbf{LDA with Non-Contextual Embeddings (Bag-Of-Words).} LDA ~\cite{blei2003latent} allows us to represent podcasts as a weighted mixture of multiple topics. The topic representation models that the same word can be used in different contexts, allowing words to appear under multiple topics. We use NLTK to tokenize the podcast transcripts and CountVectorizer\footnote{We do not use TF-IDF embeddings because TF-IDF scales down the most frequently used terms -- and \textit{discourse terms} are high-frequency terms.} to create the embeddings. We set the seed and run for a maximum of 5 iterations with a batch size of 128. We present the top 10 words for the topics, shown in Table~\ref{tab:results_topics_gender} -- in the columns \textit{Topic N} and \textit{Topic N Word List}. LDA topics are interpreted by their word list, hence, we manually assign labels to the topics in the column \textit{Topic N Categories}. \textbf{Impact of Lemmatization.} We find that lemmatization does not have a significant impact on the output (see the Appendix). Hence, for consistency \cite{reddy-etal-2021-modeling, clifton2020, yang2019more}, we do not perform lemmatization. \textbf{Impact of Topic Coherence.} We ablate the number of topics in the range $\{40, 60, 80, 100, 120, 140, 160\}$, and find that there is little variation in topic coherence (in the range of $0.394-0.428$) \cite{roder2015exploring}. Following previous work, we use $k=100$ for our experiments \cite{reddy-etal-2021-modeling, clifton2020, yang2019more}.
    \item \textbf{BERTopic with Contextual Embeddings (BERT, ChatGPT, Llama).} For BERTopic \cite{grootendorst2022bertopic},  We conduct these experiments over a subset of $10,000$ randomly-sampled podcasts. BERTopic does not require a set number of topics ahead of time -- we find that it produces approx. 50 topics for our set of 10,000 podcasts. We present the results of BERTopic on BERT embeddings, ChatGPT embeddings, and Llama embeddings in Table \ref{tab:bertopic}. \textbf{Embeddings.} For the BERT embeddings, we use the default SBERT from BERTopic. For the ChatGPT embeddings, we use the {\fontfamily{qcr}\selectfont text-embedding-3-large} model from OpenAI. For the Llama embeddings, we obtain the embeddings from {\fontfamily{qcr}\selectfont Llama-3.1-8B-Instruct} using the PromptEOL method \cite{Jiang_Huang_Luan_Wang_Zhuang_2023}. \textbf{Impact of UMAP.} UMAP is a dimensionality-reduction algorithm \cite{mcinnes2018umap} used in BERTopic to enable the dense, high-dimensional contextual embeddings to be input to HDBSCAN \cite{campello2013density}. We  experiment with values of $d=\{5,15,50,100\}$ for the reduced dimension, and find that the reduced embeddings are equivalent to each other under these different values of $d$. Hence, we conduct our experiments with the BERTopic UMAP default value of $d=5$. We hypotheize that the fine-grained information from the contextual embeddings is lost in the nearest-neighbors approximation via manifold learning from UMAP \cite{mcinnes2018umap}, hence, the output embeddings from UMAP are the same for the 3 different input embeddings.
\end{itemize}

\begin{table*}[t]
    \centering
    \resizebox{.99\textwidth}{!}{\begin{tabular}{lccccc}
    \hline
    \textbf{Topic N} & \textbf{Gender} & \textbf{$r$} & \textbf{Topic N Word List} & \textbf{Topic N Categories} & \textbf{Topic N Gender} \\
    \hline
    \multirow{2}{*}{Topic 3} &Women &\cellcolor{green!25}0.15 &\multirow{2}{*}{women, woman, men, baby, pregnant, girls, men, doctor, health, birth} &\multirow{2}{*}{Content - Pregnancy} &  \cellcolor{yellow!35} \\
                             &Men   &\cellcolor{red!25}-0.14 & & & \multirow{-2}{*}{\cellcolor{yellow!35}Women}\\ \hline
    \multirow{2}{*}{Topic 10} &Women &\cellcolor{green!25}0.10 &\multirow{2}{*}{energy, body, feel, mind, space, yoga, love, beautiful, feeling, meditation} &\multirow{2}{*}{Content - Yoga} &  \cellcolor{yellow!35}\\
                             &Men   &\cellcolor{red!25}-0.12 & & & \multirow{-2}{*}{\cellcolor{yellow!35}Women}\\ \hline
    \multirow{2}{*}{Topic 49} &Women &\cellcolor{red!25}-0.21 &\multirow{2}{*}{game, know, think, team, going, mean, play, year, one, good} &\multirow{2}{*}{Content - Sports} &  \cellcolor{orange!45} \\
                             &Men   &\cellcolor{green!25}0.17 & & & \multirow{-2}{*}{\cellcolor{orange!45}Men}\\ \hline
    \multirow{2}{*}{Topic 71} &Women &\cellcolor{green!25}0.14 &\multirow{2}{*}{christmas, sex, girl, hair, love, get, date, girls, let, wear} &\multirow{2}{*}{Content - Dating} &  \cellcolor{yellow!35} \\
                             &Men   &\cellcolor{red!25}-0.14 & & & \multirow{-2}{*}{\cellcolor{yellow!35}Women}\\ \hline
    \multirow{2}{*}{Topic 54} &Women &-- &\multirow{2}{*}{get, like, know, right, people, going, podcast, make, want, one} &\multirow{2}{*}{Discourse} & \cellcolor{orange!45} \\
                             &Men   &\cellcolor{green!25}0.12 & & & \multirow{-2}{*}{\cellcolor{orange!45}Men}\\ \hline
    \multirow{2}{*}{Topic 60} &Women &\cellcolor{red!25}-0.27 &\multirow{2}{*}{going, know, think, get, got, one, really, good, well, yeah} &\multirow{2}{*}{Discourse} & \cellcolor{orange!45}\\
                             &Men   &\cellcolor{green!25}0.20 & & & \multirow{-2}{*}{\cellcolor{orange!45}Men}\\ \hline
    \multirow{2}{*}{Topic 62} &Women &\cellcolor{green!25}0.33 &\multirow{2}{*}{like, know, really, going, people, want, think, get, things, life} &\multirow{2}{*}{Discourse} & \cellcolor{yellow!35}\\
                             &Men   &\cellcolor{red!25}-0.28 & & & \multirow{-2}{*}{\cellcolor{yellow!35}Women} \\
    \hline
    \end{tabular}}
    \caption{\textbf{LDA with Non-Contextual Embeddings (Bag-Of-Words):} The complete set of significant correlations between gender features and topic features -- \textit{both content topics and discourse topics}. Based on $r$, the Topic N Gender forms the \textbf{gendered (discourse) word lists} via Topics 54 and 60 (the masculine word lists) and Topic 62 (the feminine word list).}
    \label{tab:results_topics_gender}
\end{table*}

We examine categories of topics which occur with both the LDA and BERTopic models:\footnote{We also note a third category, \textit{language}. For details, see the Appendix.} 
\begin{itemize}
    \item \textbf{Content topics}: The content topics contain words related to content -- i.e., content related words for the topic of \textit{yoga} include \textit{energy}, \textit{body}, and \textit{meditation}. 
    \item \textbf{Discourse topics}: The discourse words contain words which \textit{are not related to content} -- including fillers (e.g., \textit{uh}, \textit{um}), discourse markers (e.g., \textit{well}, \textit{you know}, \textit{I mean}), false starts (e.g., \textit{It was, anyways, I went to Target yesterday}) and more \cite{merriam_2024, shriberg1994preliminaries}. \textbf{These words can indicate differences in the \textit{style of speech}.}  Previous works also identify discourse topics \cite{clifton2020, reddy-etal-2021-modeling, yang2019more}.
\end{itemize}

\begin{table*}[t]
    \centering
    \resizebox{.99\textwidth}{!}{\begin{tabular}{lccccc}
    \hline
    \textbf{Topic N} & \textbf{Gender} & \textbf{$r$} & \textbf{Topic N Word List} & \textbf{Topic N Categories} & \textbf{Topic N Gender} \\
    \hline
    \multirow{2}{*}{Topic 0} &Women &\cellcolor{red!25}-0.08 &\multirow{2}{*}{like, yeah, know, oh, right, podcast, got, going, think, really} &\multirow{2}{*}{Discourse} & \cellcolor{orange!45} \\
                             &Men   &\cellcolor{green!25}0.10 & & & \multirow{-2}{*}{\cellcolor{orange!45}Men}\\ \hline
    \multirow{2}{*}{Topic 2} &Women &\cellcolor{green!25}0.08 &\multirow{2}{*}{life, know, things, really, people, feel, like, want, love, going} &\multirow{2}{*}{Discourse} & \cellcolor{yellow!35}\\
                             &Men   &\cellcolor{red!25}-0.08 & & & \multirow{-2}{*}{\cellcolor{yellow!35}Women}\\ \hline
    \multirow{2}{*}{Topic 5} &Women &\cellcolor{green!25}0.08 &\multirow{2}{*}{like, know, think, yeah, episode, really, going, anchor, kind, right} &\multirow{2}{*}{Discourse} & \cellcolor{yellow!35}\\
                             &Men   &-- & & & \multirow{-2}{*}{\cellcolor{yellow!35}Women} \\
    \hline
    \end{tabular}}
    \caption{\textbf{BERTopic with Contextual Embeddings (BERT, ChatGPT, Llama):} The complete set of significant correlations between gender features and topic features for \textit{discourse topics only} (content topics are omitted).}
    \label{tab:bertopic}
\end{table*}

\begin{table*}[t]
\centering
\resizebox{0.99\textwidth}{!}{\begin{tabular}{m{1.4cm}m{1.4cm}m{0.8cm}m{5cm}m{3cm}m{5cm}m{2.9cm}}
\hline
\textbf{Topic N} &\textbf{Topic M} &\textbf{$r$} &\textbf{Topic N Word List} &\textbf{Topic N Categories} &\textbf{Topic M Word List} &\textbf{Topic M Categories}\\\hline
\multirow{2}{1.4cm}[-0.5em]{Topic 11} &Topic 54 &\cellcolor{green!25}0.11 &\multirow{2}{5cm}{data, new, technology, public, bill, theory, science, system, security, article} &\multirow{2}{3cm}{Content - Technology/ Political} &get, like, know, right, people, going, podcast, make, want, one &\cellcolor{orange!45}Discourse (Men)\\
&Topic 62 &\cellcolor{red!25}-0.20 & & &like, know, really, going, people, want, think, get, things, life &\cellcolor{yellow!35}Discourse (Women)\\ \hline
Topic 12 &Topic 54 &\cellcolor{green!25}0.24 &business, money, company, market, buy, right, million, companies, pay, sell &Content - Business&get, like, know, right, people, going, podcast, make, want, one &\cellcolor{orange!45}Discourse (Men) \\ \hline
\multirow{2}{1.4cm}[-0.5em]{Topic 79} &Topic 60 &\cellcolor{green!25}0.18 &\multirow{2}{5cm}{game, games, play, playing, like, played, nintendo, video, fun, switch} &\multirow{2}{3cm}{Content - Video Games} &going, know, think, get, got, one, really, good, well, yeah &\cellcolor{orange!45}Discourse (Men) \\
&Topic 62 &\cellcolor{red!25}-0.13 & & &like, know, really, going, people, want, think, get, things, life &\cellcolor{yellow!35}Discourse (Women) \\ \hline
\end{tabular}}
\caption{LDA with Non-Contextual Embeddings (Bag-Of-Words): Significant correlations between content topic features and \textit{gendered discourse word lists} (discourse topic features 54, 60, 62, see Table \ref{tab:results_topics_gender}) for content topic features which \textit{do not} have direct, significant correlations with gender features, but may broadly be more used by one gender.}
\label{tab:results_topics_1hop}
\end{table*}

\subsubsection{\textsc{Gender Segmenter Module.}}

We use a CNN-based model, \textit{inaSpeechSegmenter}, for state-of-the-art gender detection and segmentation~\cite{ddoukhanicassp2018}. This model allows us to analyze and approximate \textit{who}, in terms of gender, is speaking about different content topics, and in what \textit{style} they speak (discourse topics and parts-of-speech). We note that there are exceptions to this method of gender approximation, and discuss this in the Discussion-Limitations section. The model breaks the audio into time segments with five possible values for each segment: \textit{women}, \textit{men}, \textit{music}, \textit{noEnergy}, \textit{noise}. To approximate the gender makeup of each podcast, we run this model on the first 30 seconds, as podcasts often play a short snippet previewing the episode content at the beginning of the podcast and have the hosts introduce the podcast and guests, e.g., $\textit{podcast}_i$ may have a value of \textit{men seconds} $=16$, and \textit{women seconds} $=6$, and \textit{music seconds} $=8$. \textbf{\textbf{30 Second Approximation.}} We test our assumption that the first 30 seconds can approximate the gender makeup of the 10 minute versions of the podcasts. We randomly sample 100 podcasts from the 82k podcasts and run inaSpeechSegmenter \cite{ddoukhanicassp2018} on the 30 second version and 10 minute version of the podcasts; we then test for significant correlations between these versions: $r(\text{Men}\textsubscript{30 sec.},\text{Men}\textsubscript{10 min.})=0.79$ and $r(\text{Women}\textsubscript{30 sec.},\text{Women}\textsubscript{10 min.})=0.82$. Hence, the 30 second labeling is a good approximation. \textbf{French-English Language Alignment.} We test the utility of inaSpeechSegmenter for English speech gender identification. We randomly sample 10 podcasts and manually annotate the audio at the seconds-level, and find $r(\text{Men}\textsubscript{inaSpeechSegmenter},\text{Men}\textsubscript{manual})=0.995$ and $r(\text{Women}\textsubscript{inaSpeechSegmenter},\text{Women}\textsubscript{manual})=0.981$.

\subsection{Feature Correlation Measures}
We detail how we test for a significant Pearson's correlation coefficient amongst our feature vectors, $\Vec{f_i}$, which were created by our modules -- the \textsc{Conversational Parser}, the \textsc{Topic Modeler}, the \textsc{Gender Segmenter}, and an \textsc{Other} module (transcript length and duration). We test for a linear relationship between each pair of variables: $H_O: r=0$, $H_A: r\neq0$, where $H_O$ is the original hypothesis, $H_A$ is the alternate hypothesis, and $r$ is the Pearson's correlation coefficient. We follow~\citet{reddy-etal-2021-modeling} and~\citet{yang2019more} and apply a Bonferroni correction to our $\alpha$ value of 0.05, setting $\alpha=0.05/z$, where $z=\binom{124}{2}=7,626$ for LDA, representing the number of feature relationships we consider. Hence, we reject $H_O$ in favor of $H_A$ if $p\leq\alpha$. Given the largeness of $z$, our $\alpha$ value becomes small, making our criteria for significance strict and thus suitable for investigating our research questions. Furthermore, we filter our correlations $r$, such that $\|r\|>0.1$ for our LDA experiments, and $\|r\|>0.05$ for our BERTopic experiments (due to the smaller sample size of 10,000 podcasts, and fewer samples may have higher variance). Our results focus on a selection of these significant correlations; the full results are available on the project website: \url{https://www.gendered-discourse.net/extended-results}.

\section{RQ0: How Are Women and Men's Discourse Different?}

Using GDCF, our Gendered Discourse Correlation Framework shown in Figure \ref{fig:BigDiagram}, we then analyze significant correlations between between the gender features from the \textsc{Gender Segmenter} module \cite{ddoukhanicassp2018}, and the topic features from the \textsc{Topic Modeler} module \cite{blei2003latent}. We use the \textit{discourse topics} to automatically form \textit{gendered discourse word lists} via their significant correlations.

Starting with the first row of Table \ref{tab:results_topics_gender}, we see that Topic 3's word list returned by LDA with Non-Contextual Embeddings (Bag-Of-Words) (via the \textsc{Topic Modeler} module) contains the words \textit{women, woman, men, baby, pregnant, girls, men, doctor, health, birth} (in descending weighted order). Based on this word list, we manually interpret this topic as being a content topic, specifically about pregnancy, as noted in the column ``Topic N Categories.'' Then, we look to the gender correlations in the columns ``Gender'' and ``$r$,'' and see that $r(\textit{Topic 3, Women})=+0.15$ and  $r(\textit{Topic 3, Men})=-0.14$. This indicates that the topic of pregnancy positively correlates with women (identified via the \textsc{Gender Segmenter} module), and negatively correlates with men. Therefore, we associate Topic 3 (Content - Pregnancy) with Women, as noted in the ``Topic N Gender'' column. Similarly, we make these associations in the ``Topic N Gender'' column for Topics 10, 49, and 71. 

Next, we focus on the Topic 54 row. This topic is interpreted using the word list \textit{get, like, know, right, people, going, podcast, make, want, one}. This word list does not refer to any content, hence, we manually interpret this topic as being a discourse topic. Moving to the gender correlations, we see that $r(\textit{Topic 54, Women})=\emptyset$ and  $r(\textit{Topic 3, Men})=+0.12$. The reason for $r(\textit{Topic 54, Women})=\emptyset$ is because the correlation between the features \textit{Topic 54} and \textit{Women} did \textbf{not} come back as significant. However, due to the positive correlation of $0.12$ for \textit{Topic 3} and \textit{Men}, we manually associate \textit{Topic 3} with \textit{Men} in the ``Topic N Gender'' column. Similarly, we make these associations in the ``Topic N Gender'' column for Topics 60 and 62. These discourse topics, Topics 54, 60, and 62, and their top-10 word lists then, become our \textit{gendered discourse word lists}. Topics 54 and 60 are associated with men, and hence represent masculine discourse, and Topic 62 is associated with women, and hence represents feminine discourse.  We use these LDA word lists for continuity with previous work \cite{reddy-etal-2021-modeling, clifton2020, yang2019more}.

Next, looking to Table \ref{tab:bertopic}. We perform a smaller-scale (N=10,000 uniformly randomly-selected podcasts) analysis of discourse topics via BERTopic with Contextual Embeddings (BERT, ChatGPT, Llama). As stated previously, we see that the three embeddings all result in the same output topics due to the UMAP dimensionality reduction step in BERTopic, and hypothesize that this is due to the loss of fine-grained information via the nearest-neighbors approximation in UMAP \cite{mcinnes2018umap}. We see that for all three topics -- Topic 0, 2, 5 -- discourse word lists are formed and have correlations to women, men, or both, similarly to LDA with non-contextual embeddings. Hence, either method can be used depending on the individual application.

\section{RQ1: Are Discourse-Based Masculine Defaults Present in Domain-Specific Contexts?}

Using GDCF, our Gendered Discourse Correlation Framework shown in Figure \ref{fig:BigDiagram}, we analyze correlations between the gendered discourse words discovered in RQ0, and domains represented by topic features. We find that while there may not be a correlation between the gender of the speaker and the domain, there may exist discourse which is more broadly used by speakers of one gender in aggregate.

Starting with the first row of Table \ref{tab:results_topics_1hop}, we see the masculine discourse topic (Topics 54) and the feminine discourse topic (Topic 62) from RQ0 in the ``Topic M'' column. Their top-10 word lists are listed in the ``Topic M Word List'' column. Next, we see a content topic, Topic 11 in the ``Topic N'' column, and its top-10 word list in the ``Topic N Word List'' column: \textit{data, new, technology, public, bill, theory, science, system, security, article}. We manually interpret this topic, then, as being a content topic, specifically about technology/politics, and we note this in the ``Topic N Categories'' column. Then, looking to the ``$r$'' column, we see that $r(\textit{Topic 11, Topic 54})=+0.11$, and $r(\textit{Topic 11, Topic 62})=-0.20$. As Topic 54 is a masculine discourse topic, and Topic 62 is a feminine discourse topic, we conclude that the technology/political domain is somewhat dominated by masculine discourse patterns. The use of masculine discourse words in the technology/political domain constitutes a \textit{masculine default} because there is a reward associated with the masculine behavior of using certain discourse words: statistically higher salaries \cite{wagestats}. Irrespective of an individual speaker's gender, it is the \textit{use} of these masculine discourse words when discussing technology/politics which constitutes a \textit{masculine default}. These discourse words are, then, also part of the current \textit{technomasculinity} \cite{Bulut+2020}. Next, we look to the Topic 12 row. Topic 12 is a content topic which is specifically about business, and $r(\textit{Topic 12, Topic 54})=+0.24$. Since Topic 54 is a masculine discourse topic, we consider the \textit{reward} associated with the use of masculine discourse words in the business domain, and again, as business results in economic rewards, this is also a \textit{masculine default}. Finally, we look to the Topic 79 row. Topic 79 is a content topic which is specifically about video games, and $r(\textit{Topic 79, Topic 60})=+0.18$, $r(\textit{Topic 79, Topic 62})=-0.13$. Since Topic 60 is a masculine discourse topic, we consider the \textit{reward} associated with the use of masculine discourse words in the video game domain, and again, as video games are coupled with computer science \cite{Cheryan.2020, Cheryan_Plaut_Handron_Hudson_2013}, this association results in economic rewards, making this is also a \textit{masculine default}.

\section{RQ2: Are Discourse-Based Masculine Defaults Present in LLM Embeddings?}

Gender differences in LLMs are well-studied (e.g., \citet{dong-etal-2023-co2pt, Caliskan.2017, May.2019, Bolukbasi.2016}). However, masculine defaults via \textit{gender differences in discourse} in LLMs are not. As LLMs are trained in part over social media, we expect these defaults to be present in the embedding representations of \textit{gendered discourse words}.

Using D-WEAT, our Discourse Word-Embedding Association Test shown in Figure \ref{fig:embds}, we study the representation of masculine and feminine discourse words in in a state-of-the-art LLM embeddings model from OpenAI, {\fontfamily{qcr}\selectfont text-embedding-3-large}. Through hyperparameter studies, we find that the masculine discourse words have a more stable and robust representation, constituting a representational harm \cite{Blodgett.2020} and a masculine default, as this may result in better system outcomes on downstream tasks \cite{Kaneko.2021, Cao_Pruksachatkun_Chang_Gupta_Kumar_Dhamala_Galstyan_2022}. 

\subsection{D-WEAT: Discourse Word-Embedding Association Test}

We define a new intrinsic metric, D-WEAT, as an extension of WEAT \cite{Caliskan.2017} which focuses on \textit{gendered discourse words} discovered by our GDCF to ``[estimate] fairness in upstream contextualized language representation models'' \cite{Cao_Pruksachatkun_Chang_Gupta_Kumar_Dhamala_Galstyan_2022, Kaneko.2021, Bolukbasi.2016}. Similarly to WEAT, we use two sets of target words and two sets of attribute words and measure the association between them. 

\begin{itemize}
    \item \textbf{Attribute Words [$A_w, A_m$]:} We use two parallel word lists from the Word-Embedding Association Test (WEAT) 6B test \cite{Caliskan.2017} to represent the concepts of ``men'' and ``women'' in the embedding space -- see Figure \ref{fig:embds} for an illustration:
            \begin{itemize}
                \item[\labelitemiv] $A_w=$ \{\textit{women, woman, girl, she, her, sister, hers, daughter}\}
                \item[\labelitemiv] $A_m=$ \{\textit{men, man, boy, he, his, brother, him, son}\}
            \end{itemize}
    \item \textbf{Target Words [$T_w, T_m$]:} Target words are words which we ``expect to be gender neutral'' in the embedding space \cite{Kaneko.2021}. WEAT defines target words as gendered category words (e.g., \textit{math} and \textit{poetry}) to study the embedding representation of historically stereotyped subject categories (e.g., \textit{men/math} and \textit{women/poetry}) \cite{Caliskan.2017}. We define these words as gendered discourse words (e.g., \textit{going} and \textit{like} from Table \ref{tab:results_topics_gender}). This definition allows us to study the embedding representation of gendered discourse words (e.g., \textit{men/going} and \textit{women/like}). \textbf{Overlapping Words.} Our topics which represent gendered discourse, Topic 60 (masculine discourse, forms $T_m$) and Topic 62 (feminine discourse, forms $T_w$), contain some overlapping words, hence, we apply the following rules to arrive at our two lists: (i) if there is a word in the same position in both lists, it is removed, (ii) if there is a word in different positions in both lists, it remains in the list where it occurs first. Thus, we form $T_{w}$ and $T_{m}$ using our discourse topics from LDA for consistency with previous work:
        \begin{itemize}
            \item[\labelitemiv] $T_{w}=$ \{\textit{like, really, people, want, things, life, feel, time, something, right}\} These are the top weighted words \textit{post-filtering} from Topic 62 (Table \ref{tab:results_topics_gender}), which is significantly positively correlated with \textit{women} and negatively correlated with \textit{men}, representing the feminine discourse style.
            \item[\labelitemiv] $T_{m}=$ \{\textit{going, think, get, got, one, good, well, yeah, bit, week}\} These are the top 10 weighted words \textit{post-filtering} from Topic 60 (Table \ref{tab:results_topics_gender}),\footnote{Topic 54 is only positively correlated with \textit{men}, and has no significant correlation with \textit{women}; thus, we use Topic 60, as it has significant correlations for both \textit{men} ($+$) and \textit{women} ($-$).} which is significantly positively correlated with \textit{men} and negatively correlated with \textit{women}, representing the masculine discourse style.
        \end{itemize}
\end{itemize}

We also experiment with the gendered discourse topics discovered via BERTopic, and see a similar result. See the Appendix for details.

\subsubsection{Dataset Formation [$(s,s')\!\in\!S_w, \:(s,s')\!\in\!S_m$].}

Then, we form two sets of samples: $S_w$ is the \textit{women} set of samples and $S_m$ is the \textit{men} set of samples from the Spotify Podcast Dataset, as determined by $t_{women}$ and $t_{men}$, features for the total number of men or women seconds as determined by the \textsc{Gender Segmenter} module. 
For $(s,s') \in S_w$, $t_{women} \geq \tau$ and for $(s,s') \in S_m$, $t_{men} \geq \tau$, where $\tau$ is a parameter we vary to control the minimum number of men or women seconds. We search for $\tau \in \{20,25,30\}$, and study the impact of this parameter in Figures \ref{fig:tau_Sw} and \ref{fig:tau_Sm}. We form $S_w$ and $S_m$ by (i) sampling 100 podcasts which meet the $\tau$ threshold, (ii) taking 3 segments from each of those podcasts -- each of these segments is 3 sentences long,\footnote{\url{https://www.nltk.org/api/nltk.tokenize.html}.} (iii) keeping segments which have at least $\gamma$ number of words from $T_w$ for $S_w$ and $T_m$ for $S_m$. Consequently, fewer segments are kept as $\gamma$ increases. For each segment $s \in S_m$, we form $s'$ by replacing each of the words in $T_m$ with a randomly-selected word from $T_w$. For example, if:

\smallskip

$s\!=\,$\textit{And I was \textbf{going}, hey, it's cold outside...}

\smallskip

Then the word \textit{going}$\in T_m$ is replaced with a randomly-selected word from $T_w$ to form $s'$:

\smallskip

$s'\!=\,$\textit{And I was \textbf{like}, hey, it's cold outside...}

\smallskip

Hence, $\gamma=1$ for $s$ and $s'$, because there is only one word, \textit{going}, which is in $T_m$ to replace with a randomly-selected word from $T_w$. The process is similar for each sample $s \in S_w$: we form $s'$ by replacing each of the words in $T_w$ with a randomly-selected word from $T_m$. This process simulates keeping the sentence the same, except for the gendered discourse words. In future work, a more nuanced approach would involve context-aware discourse word replacement, rather than random replacement. 

\subsubsection{LLM Representation.}
We obtain the embedding representation of $s$ and $s'$ from using contextual embeddings. Specifically, the Open AI embedding model {\fontfamily{qcr}\selectfont text-embedding-3-large}\footnote{\url{https://platform.openai.com/docs/guides/embeddings}} due to the popularity of the OpenAI models. We have a similar finding with Llama embeddings, and report the results in the Appendix.
The cosine similarity between these two embedding vectors is $cos(\Vec{s},\Vec{s'}) = \Vec{s}^T \cdot \Vec{s'} / \|\Vec{s}\|\|\Vec{s'}\|$. We conventionally use the terms \textit{similarity} and \textit{distance} interchangeably -- the more similar two vectors are, the closer they are. We expect cosine similarity to be equivalent if we flip the discourse words, assuming these words are ungendered. If they move in such a way that they are more similar to either the \textit{men} or \textit{women} concepts in the embedding space, then that means that these words carry gender information.

While non-contextual embeddings were useful for topic modeling via LDA -- because LDA is designed to work on such embeddings -- they are not a good choice for this experiment. First, count embeddings are not a suitable choice because the discourse words are high-frequency, and therefore they dominate the cosine calculation and wash away the lower-frequency more informative words. Second, TF-IDF embeddings are again not a good fit for modeling discourse words because TF-IDF is specifically designed to focus on low-frequency terms via IDF scaling. Hence, we use dense contextual embeddings for our experiments to retain meaning from low-frequency words and examine the context and nuance of the high-frequency discourse words.

\subsubsection{Measuring Movement via Women and Men Percentages.}

To measure the movement in the embedding space, we calculate $\Delta_w$ and $\Delta_m$ for each $(s,s')$ pair in $S_w$ and $S_m$. The $\Delta_w$ and $\Delta_m$ indicate how $s'$ moves in relation to the \textit{women} and \textit{men} concepts, $A_w$ and $A_m$, in the embedding space, when the discourse words are replaced. Specifically, as illustrated in Figure \ref{fig:embds}, we sum the total cosine similarity between $s'$, and each of the words $w \in A_w$, and $w \in A_m$. We do the same for $s$.\footnote{We take the average over 3 calculations of the sum of the cosine similarity, to account for the small variation in the embeddings returned by the Open AI Embeddings API.} Then we calculate the movement by taking the difference of summed cosine similarity values, as shown in Equations \ref{eq:Delta_w} and \ref{eq:Delta_m}:

\begin{equation}
    \Delta_w = \sum_{a \in A_w}cos(\Vec{a},\Vec{s'}) - \sum_{a \in A_w}cos(\Vec{a},\Vec{s})
    \label{eq:Delta_w}
\end{equation}

\begin{equation}
    \Delta_m = \sum_{a \in A_m}cos(\Vec{a},\Vec{s'}) - \sum_{a \in A_m}cos(\Vec{a},\Vec{s})
    \label{eq:Delta_m}
\end{equation}

We then use two counter variables, $C_m$ and $C_w$, to indicate: How does $s'$ move in relation to $s$, and the \textit{women} and \textit{men} concepts -- $A_w$ and $A_m$? \textbf{Which concept -- women ($A_w$) or men ($A_m$) -- did $s$ move closer to, when the discourse words were replaced to form $s'$?}

\begin{equation} \label{eq:C_w}
  C_w = C_w+1 \ \left \{
  \begin{aligned}
    \text{if}\ \Delta_m, \Delta_w > 0 \ \text{and} \ \Delta_w > \Delta_m\\
    \text{if}\ \Delta_m, \Delta_w < 0 \ \text{and} \ \Delta_w < \Delta_m\\
    \text{else if}\ \Delta_w > \Delta_m\\
  \end{aligned} \right.
\end{equation} 

\begin{equation} \label{eq:C_m}
  C_m = C_m+1 \ \left \{
  \begin{aligned}
    \text{if}\ \Delta_m, \Delta_w > 0 \ \text{and} \ \Delta_m > \Delta_w\\
    \text{if}\ \Delta_m, \Delta_w < 0 \ \text{and} \ \Delta_m < \Delta_w\\
    \text{else if}\ \Delta_m > \Delta_w\\
  \end{aligned} \right.
\end{equation} 

As shown in Equations \ref{eq:C_w} and \ref{eq:C_m}, there are three possible situations in which the movement from $\Vec{s}$ to $\Vec{s'}$ can occur -- $\Vec{s'}$ moves closer to both the \textit{women} and \textit{men} concepts, $\Vec{s'}$ moves farther from both the \textit{women} and \textit{men} concepts, and $\Vec{s'}$ moves closer to one concept and farther from the other:

\begin{enumerate}
    \item For $\Delta_m, \Delta_w > 0$, $s'$ \textbf{moves closer} to both $A_w$ and $A_m$ -- the \textit{women} and \textit{men} concepts. In this case, whichever concept $s'$ moves closer to gets its corresponding counter, $C_w$ and $C_m$, incremented.

    \item For $\Delta_m, \Delta_w < 0$, $s'$ \textbf{moves farther} from both $A_w$ and $A_m$ -- the \textit{women} and \textit{men} concepts. In this case, whichever concept $s'$ moves less far from gets its corresponding counter, $C_w$ and $C_m$, incremented.

    \item For the final case, either $\Delta_m > 0$ or $\Delta_w > 0$, while the other is $< 0$, so $s'$ \textbf{moves closer} to either $A_w$ and $A_m$ \textbf{and moves farther from the other}. In this case, whichever concept $s'$ moves closer to gets its corresponding counter, $C_w$ and $C_m$, incremented.
\end{enumerate}

\begin{figure}[t] 
\centering \includegraphics[scale=0.6]{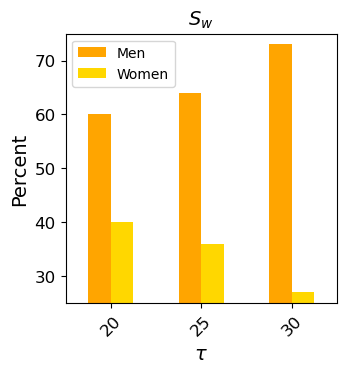} 
\caption{Impact of $\tau$ on the average percentage of $S_w$ segments which move closer to the \textit{women} concept ($A_w$) versus the \textit{men} ($A_m$) concept.} 
\label{fig:tau_Sw} 
\end{figure}

\begin{figure}[t] 
\centering \includegraphics[scale=0.6]{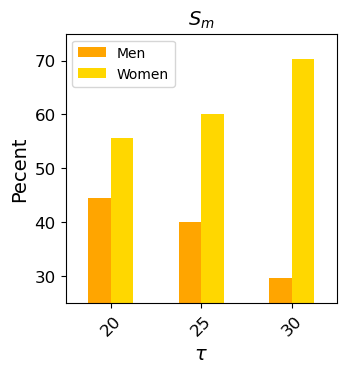} 
\caption{Impact of $\tau$ on the average percentage of $S_m$ segments which move closer to the \textit{women} concept ($A_w$) versus the \textit{men} ($A_m$) concept.} 
\label{fig:tau_Sm} 
\end{figure}

We obtain $C_w$ and $C_m$ counts and take the average. We then report these counts, $C_w$ and $C_m$, in Figures \ref{fig:tau_Sw}, \ref{fig:tau_Sm}, \ref{fig:gamma_Sw}, and \ref{fig:gamma_Sm} in terms of percentages, to normalize for the impact of $\gamma$, as there are fewer total samples which have $\geq \gamma$ discourse words to swap as $\gamma$ increases. 

\subsubsection{Impact of $\tau$.}

We study the impact of varying the minimum number of women or men seconds $\tau$ in $\{20,25,30\}$. For this analysis, we set $\gamma=6$, a middle value in our $\gamma$ study.

Starting with Figure \ref{fig:tau_Sw}, for $S_w$, we see that at $\tau=20$, the men percent is $60\%$ and the women percent is $40\%$. Moving along the $\tau$-axis, we see that the men percentage continues to increase, reaching a maximum of $\approx\!73\%$ at $\tau=30$, while the women percentage continues to decrease, reaching a minimum of $\approx\!27\%$ also at $\tau=30$. In Figure \ref{fig:tau_Sm}, for $S_m$, we see that at $\tau=20$, the men percent is $\approx\!55\%$
and the women percent is $\approx\!45\%$. Then, at $\tau=25$, the gap between the men percent and women percent widens. This gap widens again at $\tau=30$, reaching an extreme with a men percent of $\approx\!70\%$, and a women percent of $\approx\!30\%$. Comparing Figures \ref{fig:tau_Sw} and \ref{fig:tau_Sm}, we see that the initial gap at $\tau=20$ is larger on $S_w$ ($60\%$/$40\%$) versus $S_m$ ($55\%$/$45\%$). This trend continues through $\tau=30$, where the gap is larger on $S_w$ ($73\%$/$27\%$) versus $S_m$ ($70\%$/$30\%$). 

\textbf{Hence, the masculine discourse words have a more stable embedding representation than feminine discourse words (wider percentage gap on $S_w$ than $S_m$ in Figures \ref{fig:tau_Sw} and \ref{fig:tau_Sm}). This difference in embedding robustness is a representational harm \cite{Blodgett.2020}. Further, it means that an LLM can obtain better performance on downstream tasks in the presence of masculine discourse words \cite{Cao_Pruksachatkun_Chang_Gupta_Kumar_Dhamala_Galstyan_2022, Kaneko.2021} (this is a \textit{reward)} -- constituting a masculine default.}

\subsubsection{Impact of $\gamma$.}

We study the impact of varying the minimum number of swaps, $\gamma$, in the range $\{1,2,...,10\}$ for each segment for the men segments and the women segments. For this analysis, we set $\tau=30$, as this is the value for which the gap is the greatest for the men percentage and the women percentage for $S_w$ and $S_m$.

\begin{figure}[t] 
\centering \includegraphics[scale=0.55]{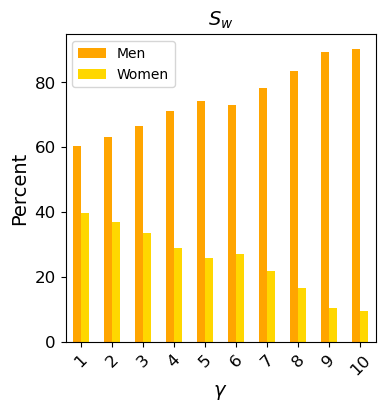} 
\caption{Impact of $\gamma$ on the average percentage of $S_w$ segments which move closer to the women concept ($A_w$) versus the men ($A_m$) concept.} 
\label{fig:gamma_Sw} 
\end{figure}

\begin{figure}[t] 
\centering \includegraphics[scale=0.55]{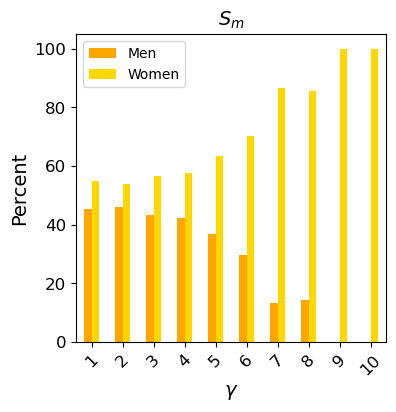} 
\caption{Impact of $\gamma$ on the average percentage of $S_m$ segments which move closer to the women concept ($A_w$) versus the men ($A_m$) concept.} 
\label{fig:gamma_Sm} 
\end{figure}

Starting with Figure \ref{fig:gamma_Sw}, at $\gamma=1$ along the $\gamma$-axis, we see that the men percent is approx. 60\%, while the women percent is approx. 40\%. This means that on average, when the discourse words from $T_w$ in each sample $s \in S_w$ were replaced with the randomly-selected discourse words from $T_m$, $\Vec{s'}$ moved closer to the man concept ($A_m$) than the women concept ($A_m$) in the contextual embedding space. As $\gamma$ increases, this gap widens, reaching an extreme at $\gamma=10$ of approx. 90\% for the men percentage and 10\% for the women percentage. This indicates that the embedding model does indeed learn gendered patterns of discourse. Similarly, as shown in Figure \ref{fig:gamma_Sm}, the gap between the men and women percent increases moving along the $\gamma$-axis, with the women percent dropping to 0\% at the extreme values of $\gamma=9,10$. 

Comparing Figures \ref{fig:gamma_Sw} and \ref{fig:gamma_Sm}, we see that the gap for $\gamma=1$ in Figure \ref{fig:gamma_Sw} is wider than the gap for $\gamma=1$ in Figure \ref{fig:gamma_Sm}, indicating that masculine discourse has a more robust representation in the embedding space. This is because with the same number of discourse word replacements, $\gamma$, more segments, on average, from the $S_w$ segments move closer to the $A_m$ man concept in the embedding space -- approx. 60\% -- while, on average, only approx. 55\% of the segments on average from the $S_m$ segments move closer to the $A_w$ women concept in the embedding space. (Note that we focus on the $\gamma=1-6$ segments as these values of $\gamma$ have lots of samples, whereas the extremes of $\gamma=7-10$ have much fewer samples and therefore higher variance, hence at these values we can interpret with less specificity that the gap between the men percentage and women percentage tends to increase.)

\textbf{The difference in embedding robustness for the masculine and feminine discourse words (wider percentage gap on $S_w$ than $S_m$ in Figures \ref{fig:gamma_Sw} and \ref{fig:gamma_Sm}) is a representational harm \cite{Blodgett.2020} which is learned by, and thus ingrained in, the widely-used embedding model, and ``such biases can easily propagate to the downstream NLP applications that use contextualised text embeddings'' \cite{Kaneko.2021, Bolukbasi.2016} (the reward). Hence, the use of masculine discourse words constitutes a masculine default.}

\section{Limitations}

Different forms of media -- such as short videos that are prevalent on TikTok, Instagram, and YouTube, long videos on YouTube, streamers on Twitch, and even text-based media such as posts on Facebook, X (Twitter), and Instagram -- may have different language patterns and styles, and further work can explore gendered discourse in these contexts. Additionally, Spotify Podcasts are produced for people who can listen to them. This imposes a socioeconomic constraint on the podcast data, as listeners must likely own an electronic device -- usually a mobile device  \cite{NPR_Research_2023}. In terms of creators, Spotify estimates that approx. 10\% of the podcasts in the dataset were made by professional creators, while the remaining 90\% were made by amateur creators \cite{clifton2020}. This long tail of amateur creators likens podcasts to social media, as Spotify for Podcasters (previously Anchor) makes it easy to produce podcasts -- i.e., producing a podcast does not necessarily require professional equipment. We lack access to certain demographic metadata in the Spotify Podcasts Dataset, including gender, age, and socioeconomic status. Our findings should not be generalized to content which is in a different format -- however, GDCF can be used to analyze this content. The Spotify data we use is limited to English speech, and $67\%$ originates from the US (as indicated by creator tags on 2,223 of the podcasts) \cite{clifton2020}.

Our study has a key limitation in that we utilize the binary gender definition,$^1$ rather than treating gender as a spectrum or otherwise modeling nonbinary genders. This lack of representation results in lowered capabilities of GDCF and D-WEAT, and hence, more nuanced approaches should be developed by future work in order to create inclusive representations of gender. Using \textit{inaSpeechSegmenter}~\cite{ddoukhanicassp2018} in the \textsc{Gender Segmenter Module}, we approximate gender via sex -- however, we note that there are many exceptions to this approximation. For example: persons with high or low voices for their sex, intersex people, transgender people, and more. While this tool, and hence this approximation of gender via sex, has been used previously (i.e., \citet{doukhan2018describing, martikainen2022exploring}), we do not claim that this is a perfect approximation, simply that it is the best one that we have at this time based on the data and model available to us towards building a discourse-based debiasing method. Future work should explore a more representative gender feature extraction step. We view this research gap as an opportunity for speech and social media researchers to create datasets to enable this technology. Rather than using binary gender, speech audio datasets can include a metadata field for self-identified genders.

\section{Theoretical Impact}

First, the use of gendered discourse words can be considered a type of \textit{gender performativity} \cite{butler1988performative, butler2009performativity, West.1987, Unger.1979, Muehlenhard.2011}, wherein the discourse words are part of a \textit{gender schema} \cite{Bem.1984, West.1987}. Hence, we identify specific words which are part of the current \textit{hegemonic masculine} strategy \cite{connell1995masculinities, connell1987gender} -- and in the domain of technology, discourse words which are part of the \textit{technomasculine} strategy \cite{Cooper_2000, lockhart2015nerd, Bulut+2020}. There exist rewards for the use of masculine discourse words in the following ways: in the domains of technology/politics and business this language is rewarded with economic rewards, and in LLMs, this language is rewarded with a more stable representation. Hence, these gendered discourse words constitute a \textit{masculine default} \cite{Cheryan.2020}, and we contribute GDCF for the discovery and analysis of gendered discourse words.

Second, D-WEAT is an intrinsic metric which can be used to debias LLMs, similarly to WEAT \cite{Caliskan.2017}, and the inclusion of discourse words broadens the debiasing task in natural language processing. We focus in this work on measuring \textit{intrinsic bias}. An important future direction includes studying gendered discourse words in the context of \textit{extrinsic bias} \cite{Blodgett.2020}, as indicated by these findings from \citet{Cao_Pruksachatkun_Chang_Gupta_Kumar_Dhamala_Galstyan_2022}:``[W]e find that correlations between intrinsic and extrinsic metrics are sensitive to alignment in notions of bias, quality of testing data, and protected groups. We also find that extrinsic metrics are sensitive to variations on experiment configurations, such as to classifiers used in computing evaluation metrics. Practitioners thus should ensure that evaluation datasets correctly probe for the notions of bias being measured.'' Hence, analyzing bias at the intrinsic and extrinsic level are two separate problems which are both important, and future work can consider extrinsic debiasing with respect to gendered discourse words.

\section{Policy Impact}

Policymakers -- in government or platforms such as Spotify -- could implement measures by which to mitigate bias in LLMs with respect to gender. Specifically, policymakers could regulate the use of D-WEAT to impose an unbiased representation of discourse words with respect to gender. D-WEAT could be run regularly, and a threshold could be set to determine what an ``acceptable'' level of bias is in a given LLM. Broadly, D-WEAT can join \textit{a set of debiasing methods, tools, and datasets} \cite{Bolukbasi.2016, Caliskan.2017, May.2019, Nangia_Vania_Bhalerao_Bowman_2020, Nadeem.2020, Guo_Yang_Abbasi_2022, He_Xia_Fellbaum_Chen_2022, Cheng_Durmus_Jurafsky_2023, dong-etal-2023-co2pt} which can be employed to regulate bias in LLMs.

\section{Ethical Impact}

A potential ethical concern is that tools used to remove bias can also be used to exacerbate bias. GDCF and D-WEAT could potentially be used to discover discourse words in audio-text corpora, and then \textit{increase} the gender bias of the LLM embeddings. This abuse of the framework would be a \textit{representational harm} \cite{Blodgett.2020}. However, a more important point is that it is hard to undo bias issues without knowing how that bias manifests; here, we provide a framework to identify and quantify this subtle gender bias so that it can be undone in powerful LLMs.

\section{Conclusion}
In this paper we analyzed \textit{discourse-based masculine defaults}. We proposed (i) the \textit{Gendered Discourse Correlation Framework (GDCF)}, a framework for identifying and analyzing gendered discourse, which we used over 15,117 podcast episodes from the Spotify Podcast Dataset \cite{clifton2020} to automatically form \textit{gendered discourse word lists}. We then proposed (ii) the \textit{Discourse Word-Embedding Association Test (D-WEAT)}, to measure the gender bias present in LLM embeddings via these \textit{gendered discourse words}, and we find that masculine discourse words have a more stable embedding representation (a representational harm), meaning an LLM can obtain better performance on downstream tasks (this is the reward). This improved embedding stability for the masculine discourse words is a masculine default. 

\section{Acknowledgments}

We would like to thank Majid Alfifi for the discussion and the help transcribing the podcasts and David Jefts for the discussion.

\bibliography{aaai22}

\appendix
\section{Appendix}

We release the code at \url{https://github.com/mariateleki/masculine-defaults} and the extended results at the project website: \url{https://www.gendered-discourse.net/extended-results}.

\begin{table}[b]
    \centering
    \caption{Means and standard deviations for number of \textit{uh}, \textit{um}, and \textit{well} tokens transcribed by WhisperX~\cite{bain2022whisperx} and Google ASR~\cite{googleSTTapi}.}
    \resizebox{.30\textwidth}{!}{\begin{tabular}{ccc}
    \hline
        \textbf{Token} & \textbf{WhisperX} & \textbf{Google ASR}  \\ 
    \hline
        \textit{uh} & \textbf{1.25} $\pm$ 2.62  & 0.10 $\pm$ 0.31 \\ 
        \textit{um} & \textbf{1.65} $\pm$ 3.03  & 0.20 $\pm$ 0.56 \\ 
        \textit{well} & \textbf{3.48} $\pm$ 2.76 & 3.51 $\pm$ 2.76 \\
    \hline
    \end{tabular}}
    \label{tab:whisperx_googleasr}
\end{table}

\subsection{WhisperX}

\subsubsection{Transcription Details with WhisperX.}
We re-transcribe the podcasts using WhisperX~\cite{bain2022whisperx}, a state-of-the-art method based on OpenAI's transformer-based Whisper ASR model, which was trained on 680,000 hours of labeled audio data~\cite{radford2023robust}. WhisperX speeds up Whisper transcription by 12x using Voice Activity Detection to detect active speech regions, and then a cut and merge strategy to allow for parallel batch transcription with Whisper. We transcribe the podcast audio using a batch size of 24. We use \textit{large-v2} for our model. The models were run across 8 machines with 60 NVIDIA RTX A4000 GPUs and 2 NVIDIA RTX A5000 GPUs. It took about a day and a half to transcribe all the podcast audio.

\subsubsection{Comparing Google ASR to WhisperX in Terms of Disfluent Token Transcription.}
As shown in Table~\ref{tab:whisperx_googleasr}, we conduct a small-scale experiment on 100 randomly-selected podcasts to evaluate the transcription quality of WhisperX as compared to Google ASR~\cite{googleSTTapi} (which was originally used to transcribe the dataset by~\citet{clifton2020}) in terms of disfluent token transcription. We fix the seeds for each random sample of size 100, and run the sampling 5 times. 

Starting with the first row of Table~\ref{tab:whisperx_googleasr}, we see the mean and standard deviation of the number of \textit{uh} tokens present in the transcripts. WhisperX transcribed on average 1.07 \textit{uh} tokens per podcast, while Google ASR transcribed on average 0.18 \textit{uh} tokens per podcast for the same podcasts. Similarly for \textit{um} and \textit{well}, the average number of \textit{um} tokens transcribed is higher for WhisperX than Google ASR. We notice similar standard deviations across WhisperX and Google ASR for the \textit{well} token, indicating consistency in the transcriptions. Thus, we hypothesize that the vast amount of training data used to train Whisper~\cite{radford2023robust} contained \textit{um} and \textit{uh} tokens, and therefore WhisperX is able to transcribe these common disfluent tokens, whilst Google ASR is less capable of transcribing these common disfluent tokens.

The large standard deviation values may be due to the heterogeneity of the podcasts, as some are scripted and likely contain less disfluent tokens, while others are unscripted and may contain many of these tokens. In the case of the \textit{uh} token, it is reasonable that the standard deviation is low for Google ASR, as there were limited \textit{uh} tokens transcribed at all (as indicated by the mean of 0.18). The case is the same for \textit{um}, which has a low standard deviation as it also has a low mean. 

\begin{table*}
    \centering
    \caption{Topic modeling with LDA: a few of the 100 topics and the top 10 weighted words for that topic.}
    \resizebox{.95\textwidth}{!}{\begin{tabular}{p{2.5cm}p{2cm}p{2cm}p{11.25cm}}
    \hline
        \textbf{Topic Number} & \textbf{Category} & \textbf{Subcategory} & \textbf{Top 10 Words for Topic} \\ 
    \hline
        Topic 5 & Content & Crime & police, crime, murder, case, killer, serial, crimes, criminal, victim, killed \\
        Topic 20 & Content & Football & jones, bowl, dallas, austin, smith, nfl, cowboys, giants, miami, eagles \\
        Topic 22 & Content & Food & coffee, drink, drinking, wine, party, tea, bar, chocolate, glass, cheese \\
        Topic 34 & Content & Medical & patients, pain, patient, disease, treatment, injury, risk, test, type, symptoms \\
        Topic 57 & Content & Church  & music, song, church, songs, album, art, mary, band, love, bible \\
        Topic 66 & Content & History & war, military, army, oil, elizabeth, russian, ii, soldiers, edward, russia \\
        Topic 70 & Content & Cars & car, drive, cars, driving, road, truck, tesla, train, traffic, miles \\
        Topic 85 & Content & Diet & food, eat, eating, weight, body, fat, day, diet, healthy, nutrition\\
        \hline
        Topic 54 & Discourse & Informal & get, like, know, right, people, going, podcast, make, want, one \\
        Topic 60 & Discourse & Informal & going, know, think, get, got, one, really, good, well, yeah\\
        Topic 62 & Discourse & Informal  & like, know, really, going, people, want, think, get, things, life \\
        Topic 88 & Discourse & Formal & one, said, would, man, see, way, says, let, say, us\\
        Topic 100 & Discourse & Informal & like, yeah, know, oh, right, got, okay, think, one, get\\
        \hline
        Topic 45 & Language & -- & tzadik, supercross, hara, mitzvot, midas, lev, tomek, barsha, attractiveness, marv\\
    \hline
    \end{tabular}}
    \label{tab:LDA_topics}
\end{table*}

\subsubsection{Non-English Podcast Episodes as Identified by WhisperX.} 
\begin{figure}[t]
\centering \includegraphics[scale=0.18]{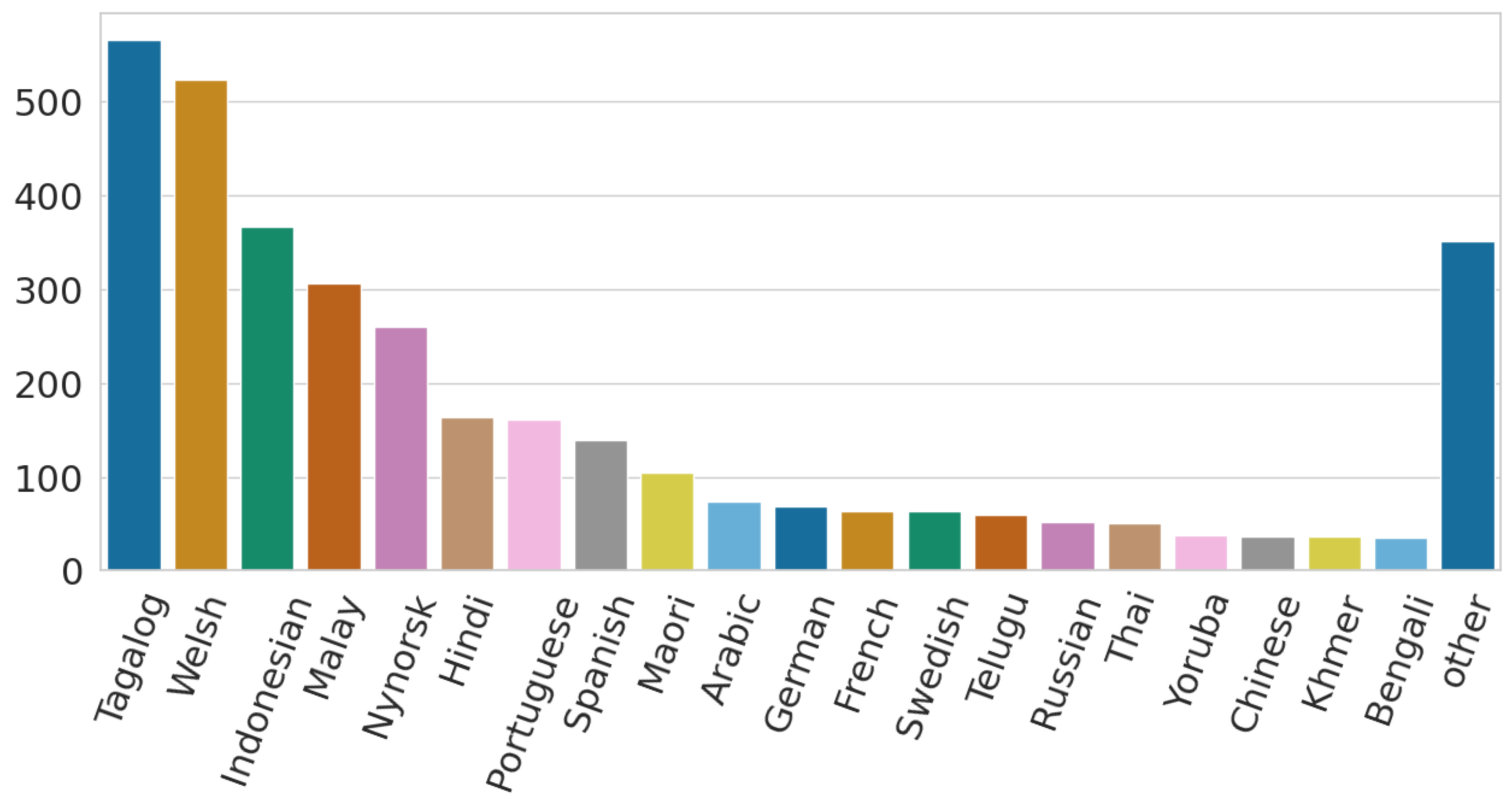} \caption{Distribution of the 3,521 non-English podcasts in the Spotify Podcasts Dataset (105,360, pre-filteration)~\cite{clifton2020} as classified by WhisperX~\cite{bain2022whisperx}. We identified a total of 63 different languages present in the corpus. See section \textit{Non-English Podcast Episodes as Identified by WhisperX} for the language makeup of the \textit{``other''} category.} 
\label{fig:nonenglish} 
\end{figure}

The Spotify 100k dataset was designed to only contain English podcasts, according to (1) the metadata, and (2) language classification based on \textit{langid.py}, a pre-trained multinomial naive Bayes learner~\cite{lui-baldwin-2011-cross,lui-baldwin-2012-langid}. However, we find that 3.34\% of the dataset (3,521 podcasts) is comprised of non-English podcasts. We show in Figure~\ref{fig:nonenglish} the distribution of the 3,521 non-English podcasts. We observe, upon looking at these previously misclassified podcasts, that they tend to be a blend of English and another language (as~\citet{clifton2020} also noted). 

In Figure~\ref{fig:nonenglish}, of the 63 non-English languages we identified, the 42 languages shown on the figure in the long-tail \textit{``other''} category are as follows: Yiddish, Hungarian, Dutch, Latin, Tamil, Korean, Malayalam, Urdu, Lithuanian, Javanese, Romanian, Latvian, Hebrew, Swahili, Myanmar, Vietnamese, Galician, Marathi, Afrikaans, Japanese, Norwegian, Turkish, Greek, Nepali, Shona, Finnish, Bulgarian, Sinhala, Sanskrit, Italian, Slovenian, Kannada, Breton, Punjabi, Gujarati, Haitian Creole, Hawaiian, Polish, Danish, Persian, Estonian, and Amharic.

In the LDA topics, we note a third category of language-related topics -- in addition to the \textit{content} and \textit{discourse} categories -- \textit{language}. For example, Topic 45 contains words of multiple different languages: ``tzadik'' and ``mitzvot'' are Hebrew words, ``hara'' is a Hindi word, ``lev'' is a Bulgarian word, and ``attractiveness'' is an English word. We suspect that episodes which are highly weighted for this topic are primarily English -- as they passed the WhisperX English language filter -- interwoven with words of other languages. Examples of topics are shown in Table~\ref{tab:LDA_topics}

\subsubsection{The Importance of High-Quality Transcription.}
The podcasts were originally transcribed using Google Automatic Speech Recognition (ASR)~\cite{googleSTTapi, clifton2020}. Per~\citet{clifton2020}, Google ASR had a sample word error rate (WER) of 18.1\% on the podcasts. Comparably, WhisperX shows a WER of 13.8\% on the Switchboard dataset~\cite{swb_whisperx_eval}, which is also a conversational dataset. Google ASR tends to transcribe less disfluent discourse tokens -- such as \textit{uh}, \textit{um}, and \textit{well} -- than WhisperX, as shown in Table~\ref{tab:whisperx_googleasr}.\footnote{\citet{clifton2020} notes: \textit{``We also anticipate that the state of the art in automatic speech recognition will improve in the coming years, allowing for more accurate automatic transcriptions.''}}

\begin{table*}
\centering
\caption{Code and licenses.}
\resizebox{\textwidth}{!}{\begin{tabular}{llp{4.5cm}} \hline
 & \textbf{Link} & \textbf{License} \\ \hline
The Spotify Podcast Dataset &  \url{https://podcastsdataset.byspotify.com/} & Creative Commons Attribution 4.0 International License \\
WhisperX & \url{https://github.com/m-bain/whisperX} & BSD-4-Clause License \\
CountVectorizer & \url{https://scikit-learn.org/stable/modules/generated/sklearn.feature_extraction.text.CountVectorizer.html} & BSD License \\
LatentDirichletAllocation & \url{https://scikit-learn.org/stable/modules/generated/sklearn.decomposition.LatentDirichletAllocation.html} & BSD License \\
NLTK & \url{https://www.nltk.org/howto/corpus.html?highlight=stopwords} & Apache License 2.0 \\
\textit{inaSpeechSegmenter} & \url{https://github.com/ina-foss/inaSpeechSegmenter} & The MIT License \\
\textit{english-fisher-annotator} & \url{https://github.com/pariajm/english-fisher-annotations} & None    \\ 
    \hline
    \end{tabular}}
    \label{tab:license}
\end{table*}

These disfluencies, as well as improved overall transcription quality, support more fine-grained analysis of how people communicate. Indeed, ``[s]pontaneous human speech is notoriously disfluent''~\cite{brennan2001listeners}. While some podcasts are scripted, and thus match the traditional training data of ASR systems -- audiobooks --~\cite{panayotov2015librispeech}, many are not, and are instead unscripted, and consist of spontaneous, conversational, disfluent speech (e.g., interviews, talk shows, etc.). People ``are highly
sensitive to hesitation disfluencies in speech''~\cite{corley2008hesitation}, as ``words preceded by disfluency [are] more likely to be remembered'' by listeners~\cite{corley2007s}. The different types of disfluency even produce different levels of \textit{memory boosts}~\cite{diachek2023effect}. This is important, as disfluencies are common, occurring at a rate of approx. 4-6 disfluent words per 100 words~\cite{tree1995effects, branigan1999non}. On the speaker side, ``disfluencies are associated with an increase in planning difficulty''~\cite{bortfeld2001disfluency}. ``[S]peakers use uh and um to announce that they are initiating what they expect to be a minor (uh), or major (um), delay in speaking''~\cite{clark2002using}, and this makes sense, as ``[disfluencies] precede relatively unpredictable lexical items [and] relatively infrequent lexical items''~\cite{beattie1979contextual}. Hence, towards characterizing the podcast content, we aim to retain disfluencies in our audio transcriptions using WhisperX.

Disfluencies are also interesting in the context of gender, where ``filled pauses may serve to `hold the floor'''~\cite{shriberg1996disfluencies} -- i.e., to verbally take up more time in conversation. Additionally, there is a difference in disfluent, filler speech production with respect to gender~\cite{bortfeld2001disfluency}. The podcasts, being spoken content and representing human conversational speech, thus provide a new opportunity to study gendered speech differences. 

\subsection{Large-Scale Confirmation of Small-Scale Studies}
\textbf{\textsc{Other} Module Details.}
\textit{Duration} refers to the floating-point time in minutes per episode. This feature was provided as part of the Spotify Podcasts Dataset~\cite{clifton2020}. \textit{Speech Rate} is approximated by measuring the integer number of words in each 10-minute truncated WhisperX transcript. A transcript with a higher word count has more words spoken in the same amount of time (10 minutes) as a transcript with a lower word count.

\begin{table}
    \centering \small
    \caption{Significant correlations between duration, speech rate, and gender.}
    \begin{tabular}{lcc}
    \hline
                       & \textbf{Women} & \textbf{Men} \\ \hline
    Duration           & \cellcolor{red!25}-0.17  & \cellcolor{green!25}0.12 \\
    Speech Rate        & --     & \cellcolor{green!25}0.15 \\\hline
    \end{tabular}
    \label{tab:duration-gender}
\end{table}

\begin{table*}[t]
    \centering
    \caption{Significant correlations between content topics and content topics.}
\resizebox{.99\textwidth}{!}{\begin{tabular}{m{1.25cm}m{1.25cm}m{0.6cm}m{5.75cm}m{3cm}m{5.75cm}m{3cm}}
\hline
\textbf{Topic N} &\textbf{Topic M} &\textbf{$r$} &\textbf{Topic N Word List} &\textbf{Topic N Categories} &\textbf{Topic M Word List} &\textbf{Topic M Categories} \\ \hline
Topic 3 &Topic 34 &\cellcolor{green!25}0.10 &women, woman, men, baby, pregnant, girls, men, doctor, health, birth &Content - Pregnancy &patients, pain, patient, disease, treatment, injury, risk, test, type, symptoms &Content - Medical \\ \hline
Topic 5 &Topic 53 &\cellcolor{green!25}0.31 &police, crime, murder, case, killer, serial, crimes, criminal, victim, killed &Content - Crime &would, family, years, could, children, father, life, time, home, young &Content - Family \\ \hline
\multirow{2}{1.25cm}[-0.5em]{Topic 11} &Topic 63 &\cellcolor{green!25}0.34 &\multirow{2}{5.75cm}{data, new, technology, public, bill, theory, science, system, security, article} &\multirow{2}{4.2cm}{Content - Technology/ Politics} &people, world, black, country, america, states, history, white, american, united &Content - USA \\
&Topic 66 &\cellcolor{green!25}0.12 & & &war, military, army, oil, elizabeth, russian, ii, soldiers, edward, russia &Content - European History \\  \hline
Topic 63 &Topic 66 &\cellcolor{green!25}0.26 &people, world, black, country, america, states, history, white, american, united &Content - USA &war, military, army, oil, elizabeth, russian, ii, soldiers, edward, russia &Content - European History \\ \hline
Topic 12 &Topic 72 &\cellcolor{green!25}0.14 &business, money, company, market, buy, right, million, companies, pay, sell &Content - Business &bitcoin, adam, people, crypto, show, coin, network, mining, coins, meister &Content - Cryptocurrency \\ \hline
Topic 13 &Topic 80 &\cellcolor{green!25}0.33 &club, soccer, la, phil, well, mexico, de, real, lucas, madrid &Content - Soccer &goal, goals, league, season, cup, yeah, points, obviously, chelsea, premier &Content - Soccer \\ \hline
\multirow{3}{1.5cm}[-1.15em]{Topic 49} &Topic 13 &\cellcolor{green!25}0.13 &\multirow{3}{5.75cm}{game, know, think, team, going, mean, play, year, one, good} &\multirow{3}{4.2cm}{Content - Sports} &club, soccer, la, phil, well, mexico, de, real, lucas, madrid &Content - Soccer \\ 
&Topic 20 &\cellcolor{green!25}0.41 & & &jones, bowl, dallas, austin, smith, nfl, cowboys, giants, miami, eagles &Content - Football \\
&Topic 25 &\cellcolor{green!25}0.30 & & &state, florida, college, texas, south, north, carolina, michigan, georgia, ohio &Content - USA States \\ \hline
Topic 22 &Topic 71 &\cellcolor{green!25}0.13 &coffee, drink, drinking, wine, party, tea, bar, chocolate, glass, cheese &Content - Food/Drink &christmas, sex, girl, hair, love, get, date, girls, let, wear &Content - Dating \\ \hline
\end{tabular}}
    \label{tab:content-topics_content-topics}
\end{table*}

\begin{table}
    \centering \small
    \caption{Significant correlations between POS and gender.}
\begin{tabular}{lcc}
\hline
\textbf{Part-of-Speech}                   & \textbf{Women} & \textbf{Men} \\ \hline
Edited & \cellcolor{red!25}-0.14  & \cellcolor{green!25}0.16 \\
Parenthetical    & \cellcolor{red!25}-0.11  & \cellcolor{green!25}0.12 \\
Adjective Phrase   & \cellcolor{green!25}0.13   & --    \\
Noun Phrase     & --      & \cellcolor{green!25}0.17 \\
Prepositional Phrase     & \cellcolor{red!25}-0.1   & \cellcolor{green!25}0.14 \\
\hline
\end{tabular}
    \label{tab:POS_gender}
\end{table}

\begin{table}[t]
    \centering
    \caption{Significant correlations between parts-of-speech, and formal and informal discourse (as determined by their top weighted words).}
    \resizebox{0.85\columnwidth}{!}{\begin{tabular}{lcc}
    \hline
    \textbf{Part-of-Speech} & \textbf{Topic 100:}           & \textbf{Topic 88:} \\
                        & \textbf{Discourse -}          & \textbf{Discourse -} \\
                        & \textbf{Informal}             & \textbf{Formal} \\
    \hline
    Interjection       &\cellcolor{green!25}0.86 &\cellcolor{red!25}-0.20 \\
    Edited     &\cellcolor{green!25}0.37 &\cellcolor{red!25}-0.22 \\
    Parenthetical       &\cellcolor{green!25}0.17 &\cellcolor{red!25}-0.22 \\
    Adjective Phrase      &\cellcolor{green!25}0.17 &\cellcolor{red!25}-0.19 \\
    Adverb Phrase       &\cellcolor{green!25}0.22 &\cellcolor{red!25}-0.30 \\
    Noun Phrase        &\cellcolor{green!25}0.14 &\cellcolor{red!25}-0.12 \\
    Prepositional Phrase  &\cellcolor{red!25}-0.43  &- \\
    Simple Declarative Clause &\cellcolor{green!25}0.19 &\cellcolor{red!25}-0.18 \\
    Verb Phrase         &\cellcolor{green!25}0.17 &\cellcolor{red!25}-0.18 \\
    \hline
    \end{tabular}}
    \label{tab:POS_wormal-topics_informal-topics}
\end{table}

\textbf{\textsc{Conversational Parser} Module Details.}
We use a state-of-the-art parsing model, \textit{english-fisher-annotator}~\citet{jamshid-lou-johnson-2020-improving}, for parsing sentences and obtaining part-of-speech (POS) counts for each 10-minute transcript. As we used WhisperX for the transcriptions, we have high-quality transcripts -- which also include more disfluencies (i.e., \textit{um}, \textit{uh}, and more), see Figure \ref{tab:whisperx_googleasr} -- and therefore use this parsing model, which was designed for use in disfluent, conversational settings. 

\begin{table*}
    \centering
    \caption{Significant correlations between discourse topics. Gender correlation labels for Topics N and M are assigned based on significant correlations from Table~\ref{tab:results_topics_gender}.}
\resizebox{.90\textwidth}{!}{\begin{tabular}{m{1.25cm}m{1.25cm}m{0.75cm}m{5.75cm}m{3cm}m{5.75cm}m{3cm}}
\hline
\textbf{Topic N} &\textbf{Topic M} &\textbf{$r$} &\textbf{Topic N Word List} &\textbf{Topic N Categories} &\textbf{Topic M Word List} &\textbf{Topic M Categories} \\ 
\hline
Topic 60 &Topic 62 &\cellcolor{red!25}-0.34 &going, know, think, get, got, one, really, good, well, yeah &\cellcolor{orange!45}Discourse - Informal (Men) & like, know, really, going, people, want, think, get, things, life &\cellcolor{yellow!35}Discourse - Informal (Women) \\ \hline
\multirow{3}{1.4cm}[-1.15em]{Topic 100} &Topic 60 &\cellcolor{red!25}-0.13 &\multirow{3}{5.75cm}{like, yeah, know, oh, right, got, okay, think, one, get} &\multirow{3}{3cm} {Discourse - Informal} &going, know, think, get, got, one, really, good, well, yeah       &\cellcolor{orange!45}Discourse - Informal (Men) \\  
&Topic 62 &\cellcolor{red!25}-0.29 & & &like, know, really, going, people, want, think, get, things, life &\cellcolor{yellow!35}Discourse - Informal (Women) \\ 
&Topic 88 &\cellcolor{red!25}-0.12 & & &one, said, would, man, see, way, says, let, say, us &Discourse - Formal \\
\hline
\end{tabular}}
    \label{tab:discourse-topics_discourse-topics}
\end{table*}

The parsing model, \textit{english-fisher-annotator}, specializes in handling annotation of the edited part-of-speech nodes, which often arises in conversational, spontaneous speech. The model was evaluated on the Switchboard dataset~\cite{godfrey1997switchboard, mitchell1999treebank}, and is scored based on its performance on the disfluent edited, interjection, and parenthetical node types. In the sentence fragment \textit{It was cold, oh I think, it was hot outside...}, \textit{It was cold} is an edited node, \textit{oh} is an interjection, and \textit{I think} is a parenthetical; deleting these from the sentence would form a fluent sentence, hence these are the disfluent node types. The model scores $P=92.5$, $R=97.2$, and $F=94.8$ for edited, interjection, and parenthetical nodes~\cite{jamshid-lou-johnson-2020-improving}. 

We annotated all of the truncated transcripts to obtain their parse trees on a sentence-level. We ran \textit{english-fisher-annotator} on a single machine with 2 NVIDIA TITAN Xp GPUs. We obtain the counts for each POS by counting the number of times that label occurs across all the parse trees for each 10-minute transcript. We obtain the parse trees on a sentence level, and truncate all sentences in each transcript to 300 tokens for compatibility with \textit{english-fisher-annotator}. We use a subset of all possible POS labels (see Figure \ref{tab:POS_wormal-topics_informal-topics}) for our analysis~\cite{taylor2003penn, jamshid-lou-johnson-2020-improving}.

\citet{taylor2003penn} describe the creation of the Penn Treebank-3 dataset for evaluating the models.~\citet{charniak2001edit} formalize the evaluation metrics for the parsing-based disfluency annotation task. Disfluency parsing is an established line of work~\cite{johnson2004tag, honnibal2014joint, tran2017parsing, lou2019neural, jamshid-lou-johnson-2020-improving}. 

As shown in Equation~\ref{eq:span-labeling}, the model classifies all of the spans in a string, from position $i$ to position $j$ with a label $l$ based on the classification scores for each span. The model then calculates the score for each parse tree, $s(T)$, by summing $s(i,j,l)$.

\begin{equation} \label{eq:span-labeling}
s(T)= \sum_{(i,j,l) \in T}^{} s(i,j,l)
\end{equation}

Then, out of all the possible parse trees,  $s(T)$, the highest-scoring parse tree, $\hat{T}$, is selected as the parse tree for that sentence, as shown in Equation~\ref{eq:highest-scoring-parse-tree}.

\begin{equation} \label{eq:highest-scoring-parse-tree}
\hat{T} = \underset{T}{\mathrm{argmax}}\; s(T)
\end{equation}

We use the \textit{swbd\_fisher\_bert\_Edev.0.9078} model checkpoint.

\subsubsection{How are gender, duration, and speech rate related?}
Starting with the first row in Table~\ref{tab:duration-gender}, we see that \textit{Duration} and \textit{women} have a negative correlation ($-0.17$), and \textit{Duration} and \textit{men} have a positive correlation ($0.12$). This indicates that the \textbf{more} minutes in duration a podcast episode is, there tends to be, then, \textbf{more} seconds of men speech in the first 30 seconds of the podcast episode. Conversely, for the case of \textit{Duration} and the feature \textit{Gender - women}, the correlation is $-0.17$, indicating that the \textbf{more} minutes in duration the podcast is, there tends to be \textbf{less} seconds of women speech in the first 30 seconds of the podcast episode.

In the second row, we see that \textit{Speech Rate} and \textit{women} do not have a significant correlation, whilst \textit{Speech Rate} and \textit{men} have a significant positive correlation ($0.15$). This indicates that the more masculine a podcast is, the faster that the rate of speech is in that podcast.

These findings are consistent with~\citet{leaper2007meta} and~\citet{james1993understanding}, who also find that men are overall more ``talkative'' than women, which relates to men ``hold[ing] the floor''~\cite{shriberg1996disfluencies}.~\citet{shriberg1996disfluencies} notes that this may be due to gender being confounded with other variables -- such as education level and occupation -- that men are able to ``hold the floor'' longer than women.

\subsubsection{Which topics are related?}
In Table~\ref{tab:content-topics_content-topics}, we can see that similar topics have positive correlations with each other. Topics 3 (Pregnancy) and 34 (Medical) are positively correlated. Topics 5 (Crime) and 53 (Family) are correlated. Topics 11 (Technology/Politics), 63 (USA), and 66 (European History) are all positively correlated with each other. Topics 12 (Business) and 72 (Cryptocurrency) are positively correlated. Topic 49 (Sports) is positively correlated with Topics 13 (Soccer), 20 (Football), and 25 (USA States). Topics 22 (Food/Drink) and 71 (Dating) are positively correlated. These correlations imply that there may be some clustering structure to the topics, and deserves further study.

\subsubsection{How do parts-of-speech vary by gender?}
In Table~\ref{tab:POS_gender}, disfluent parts-of-speech (edited and parenthetical~\cite{jamshid-lou-johnson-2020-improving}) are negatively correlated with \textit{women}, and positively correlated with \textit{men}, indicating that men tend to be more disfluent in their speech. \textit{Adjective Phrases} are positively correlated with \textit{women}, while \textit{Noun Phrases} and \textit{Prepositional Phrases} are positively correlated with \textit{men}. This indicates that women tend to use more descriptive words (adjective phrases) in their speech.~\citet{shriberg1996disfluencies} observed an association between increased filled pauses and men, and states that ``filled pauses may serve to `hold the floor,''' but also that gender is confounded with other variables such as education level and occupation.~\citet{bortfeld2001disfluency} also found that men produced more fillers than women in their speech.

\subsubsection{How do parts-of-speech vary for informal and formal discourse?}

\begin{table*}
    \centering
    \caption{\textbf{LDA with Non-Contextual Embeddings (Bag-Of-Words) \textit{with Lemmatization}}: Significant correlations between gender features and topic features for \textit{discourse topics only} (content topics are omitted).}
    \resizebox{.99\textwidth}{!}{\begin{tabular}{lccccc}
    \hline
    \textbf{Topic N} & \textbf{Gender} & \textbf{$r$} & \textbf{Topic N Word List} & \textbf{Topic N Categories} & \textbf{Topic N Gender} \\
    \hline
    \multirow{2}{*}{Topic 30} &Women &\cellcolor{green!25}0.17 &\multirow{2}{*}{like, know, yeah, go, get, think, really, say, want, right} &\multirow{2}{*}{Discourse} & \cellcolor{yellow!35} \\
                             &Men   &\cellcolor{red!25}-0.12 & & & \multirow{-2}{*}{\cellcolor{yellow!35}Women}\\ \hline
    \multirow{2}{*}{Topic 75} &Women &\cellcolor{red!25}-0.31 &\multirow{2}{*}{get, go, know, think, yeah, game, year, play, good, one} &\multirow{2}{*}{Discourse} & \cellcolor{orange!45}\\
                             &Men   &\cellcolor{green!25}0.25 & & & \multirow{-2}{*}{\cellcolor{orange!45}Men}\\ \hline
    \multirow{2}{*}{Topic 91} &Women &\cellcolor{green!25}0.13 &\multirow{2}{*}{people, go, thing, really, know, get, want, think, work, time} &\multirow{2}{*}{Discourse} & \cellcolor{yellow!35}\\
                             &Men   &\cellcolor{red!25}-0.11 & & & \multirow{-2}{*}{\cellcolor{yellow!35}Women} \\
    \hline
    \end{tabular}}
    \label{tab:lemmatization_true}
\end{table*}

\begin{table*}
    \centering
    \caption{\textbf{LDA with Non-Contextual Embeddings (Bag-Of-Words) \textit{without Lemmatization}}: Significant correlations between gender features and topic features for \textit{discourse topics only} (content topics are omitted).}
    \resizebox{.99\textwidth}{!}{\begin{tabular}{lccccc}
    \hline
    \textbf{Topic N} & \textbf{Gender} & \textbf{$r$} & \textbf{Topic N Word List} & \textbf{Topic N Categories} & \textbf{Topic N Gender} \\
    \hline
    \multirow{2}{*}{Topic 45} &Women &\cellcolor{green!25}0.29 &\multirow{2}{*}{know, like, really, people, going, think, want, things, get, kind} &\multirow{2}{*}{Discourse} & \cellcolor{yellow!35} \\
                             &Men   &\cellcolor{red!25}-0.23 & & & \multirow{-2}{*}{\cellcolor{yellow!35}Women}\\ \hline
    \multirow{2}{*}{Topic 74} &Women &\cellcolor{red!25}-0.25 &\multirow{2}{*}{think, know, going, game, got, team, yeah, good, year, one} &\multirow{2}{*}{Discourse} & \cellcolor{orange!45}\\
                             &Men   &\cellcolor{green!25}0.20 & & & \multirow{-2}{*}{\cellcolor{orange!45}Men}\\ \hline
    \multirow{2}{*}{Topic 95} &Women &\cellcolor{red!25}-0.12 &\multirow{2}{*}{one, going, well, get, got, would, time, yeah, back, go} &\multirow{2}{*}{Discourse} & \cellcolor{orange!45}\\
                             &Men   &\cellcolor{green!25}0.06 & & & \multirow{-2}{*}{\cellcolor{orange!45}Men} \\
    \hline
    \end{tabular}}
    \label{tab:lemmatization_false}
\end{table*}

In Table~\ref{tab:POS_wormal-topics_informal-topics}, we see that \textit{Topic 100}, informal language, tends to have more disfluencies (\textit{Interjection}, \textit{Edited}, and \textit{Parenthetical} parts-of-speech~\cite{jamshid-lou-johnson-2020-improving}) than \textit{Topic 88}, formal language, as shown in Table~\ref{tab:results_topics_gender}. 

\textit{Topic 100}, informal language, is highly correlated (0.86) with increased \textit{Adjective Phrases}, \textit{Adverb Phrases}, and \textit{Noun Phrases}, while \textit{Topic 88}, formal language, is \textit{not} correlated with these parts-of-speech. \textit{Topic 100}, informal language, is also negatively correlated with \textit{Prepositional Phrases}. This means that informal language tends to be more descriptive, as characterized by more adjective phrases, adverb phrases, and noun phrases. 

\subsubsection{Are discussion styles distinct?}
Table~\ref{tab:discourse-topics_discourse-topics} examines the relationship between informal and formal discourse topics. Starting with the first row, the correlation between \textit{Topic 60} and \textit{Topic 62} is -0.34. \textit{Topic 62} is a topic which is characterized by informal speech, as shown by its word list, and from Table~\ref{tab:results_topics_gender}, \textit{Topic 62} has a positive correlation with \textit{women} (0.33) and a negative relationship with \textit{men} (-0.28), making it a \textit{women} topic. This contrasts with \textit{Topic 60}, which is primarily a \textit{men} topic. Hence, the data indicates that men informal language and women informal language tend not to co-occur. This aligns with the correlation value for the \textit{men} and \textit{women} features: -0.76. \textit{Topic 100} is extremely informal, as it includes swear words. We see that it is distinct from the other informal discourse topics, \textit{Topic 60} and \textit{Topic 62}, as it has -0.13 and -0.29 correlation values with these topics.

\subsection{Impact of Lemmatization on LDA with Non-Contextual Embeddings}

We find that lemmatization before the creation of the non-contextual bag-of-words embeddings does not have a significant impact on the quality of the discourse topics created by LDA. Looking to Tables \ref{tab:lemmatization_true} and \ref{tab:lemmatization_false}, we see that discourse topics are still formed with and without the lemmatization step, and that these topics have significant correlations with women and men.

While the topics formed with lemmatization are different than the topics formed without lemmatization, we still see patterns of gendered speech that emerge within these discourse topics, as indicated by $r$. Even on a smaller sample size of 10,000 podcasts, many of the correlations are high ($\lvert r \rvert > 0.20$).

\subsection{WEAT versus SEAT}
WEAT \cite{Caliskan.2017} words are advantageous as compared to SEAT \cite{May.2019} sentences, because ``the context is artificial, which does not reflect the natural usage of a word.'' \cite{Nadeem.2020}.

\subsection{D-WEAT with BERTopic Discourse Topics}

We run the D-WEAT experiment with the discourse topics formed via BERTopic with contextual embeddings (BERT, ChatGPT, Llama). \textbf{Target Words [$T_w, T_m$].} We form $T_{w}$ and $T_{m}$ in the same way, using our discourse topics from BERTopic (see Table \ref{tab:bertopic}):
\begin{itemize}
    \item[\labelitemiv] $T_{w}=$ \{\textit{life, know, things, really, people, feel, want, love, way, person}\} These are the top weighted words \textit{post-filtering} from Topic 2,\footnote{Topic 5 is only positively correlated with \textit{women}, and has no significant correlation with \textit{women}; thus, we use Topic 0, as it has significant correlations for both \textit{women} ($+$) and \textit{men} ($-$).} which is significantly positively correlated with \textit{women} and negatively correlated with \textit{men}, representing the feminine discourse style.
    \item[\labelitemiv] $T_{m}=$ \{\textit{like, yeah, oh, right, podcast, got, going, think, okay, f***ing}\} We censor the last word for presentation, but not in the experiment. These are the top 10 weighted words \textit{post-filtering} from Topic 0, which is significantly positively correlated with \textit{men} and negatively correlated with \textit{women}, representing the masculine discourse style.
\end{itemize}

\textbf{LLM Representation.} We use {\fontfamily{qcr}\selectfont text-embedding-3-large} from Open AI via the API with a single call. \textbf{Impact of $T_{w}$ and $T_{m}$.} We study the impact of varying $T_{w}$ and $T_{m}$ formed via BERTopic with contextual embeddings (BERT, ChatGPT, Llama). We fix $\tau = 30.0$ and $\gamma = 6$. We make 1 API call. We use 1 seed. We find that for $S_w$, the men percentage is $100\%$ and the women percentage is $0\%$. For $S_m$, the men percentage is approx. $22\%$ and the women percentage is approx. $78\%$. This finding indicates that the gendered discourse words discovered via BERTopic are represented in a gender-imbalanced way in the embedding model, and hence, this is a masculine default.

\subsection{D-WEAT with Llama Embeddings}

We run the D-WEAT experiment with Llama embeddings for the LLM representation. We obtain the embeddings from {\fontfamily{qcr}\selectfont Llama-3.1-8B-Instruct} using the PromptEOL method \cite{Jiang_Huang_Luan_Wang_Zhuang_2023}. \textbf{Impact of Embeddings.} We study the impact of varying the embedding model. We fix $\tau = 30.0$ and $\gamma=6$. We make 1 API call. We use 1 seed. We find that for $S_w$, the men percentage is $70\%$ and the women percentage is $30\%$. For $S_m$, the men percentage is $50\%$ and the women percentage is $50\%$. This finding indicates that the gendered discourse words discovered via BERTopic are represented in a gender-imbalanced way in the embedding model -- in that men obtain a stable representation (no gap between women and men percentages) while women do not (larger gap between women and men percentages). Hence, this is a masculine default.

\end{document}